\documentclass[twoside,11pt]{article}
\usepackage{journal, rawfonts}
\usepackage{algorithm}
\usepackage{algorithmic}
\usepackage{cite}
\usepackage{epigraph}
\usepackage{amsmath,amssymb,amsfonts, pifont, subcaption} 
\newcommand{\cmark}{\ding{51}}%
\newcommand{\xmark}{\ding{55}}%
\usepackage{makecell}
\usepackage{enumitem}
\usepackage{booktabs} 
\usepackage{graphicx}
\DeclareMathOperator*{\argmin}{arg\,min}

\usepackage{textcomp}

\newcommand\blfootnote[1]{%
  \begingroup
  \renewcommand\thefootnote{}\footnote{#1}%
  \addtocounter{footnote}{-1}%
  \endgroup
}

\def\BibTeX{{\rm B\kern-.05em{\sc i\kern-.025em b}\kern-.08em
    T\kern-.1667em\lower.7ex\hbox{E}\kern-.125emX}}
    
\begin{document}
\title{\textbf{SCouT}: \textbf{S}ynthetic \textbf{Cou}nterfactuals via Spatiotemporal \textbf{T}ransformers for Actionable Healthcare}

\jairheading{1}{2022}{1-32}{--/--}{--/--}
\ShortHeadings{SCouT: Synthetic Counterfactuals via Spatiotemporal Transformers for Actionable Healthcare}
{ Dedhia, Balasubramanian, \& Jha}
\firstpageno{1}

\author{\name Bhishma Dedhia$^\dagger$ \email bdedhia@princeton.edu \\
       \addr Dept. of Electrical \& Computer Engineering, Princeton University \\
       Princeton, NJ 08540 USA
       \AND
       \name Roshini Balasubramanian$^\dagger$ $^\ddagger$ \email  roshinib@princeton.edu\\
       \addr Dept. of Operations Research and Financial Engineering, Princeton University \\ Princeton, NJ 08540, USA 
       \AND
       \name Niraj K. Jha \email jha@princeton.edu \\
       \addr Dept. of Electrical \& Computer Engineering, Princeton University \\
       Princeton, NJ 08540 USA}
\maketitle

\begin{abstract}
The Synthetic Control method has pioneered a class of powerful data-driven techniques to
estimate the counterfactual reality of a unit from donor units. At
its core, the technique involves a linear model fitted on the pre-intervention period that
combines donor outcomes to yield the counterfactual. However, linearly combining spatial information at each time
instance using time-agnostic weights fails to capture important inter-unit and
intra-unit temporal contexts and complex nonlinear dynamics of real data. We instead propose an approach to use local spatiotemporal
information before the onset of the intervention as a promising way to estimate the
counterfactual sequence. To this end, we suggest a Transformer model that leverages
particular positional embeddings, a modified decoder attention mask, and a novel pre-training task to perform
spatiotemporal sequence-to-sequence modeling.  Our experiments on synthetic data demonstrate the efficacy of
our method in the typical small donor pool setting  and its robustness against noise. We also generate
actionable healthcare insights at the population and patient levels by simulating a state-wide public health policy to evaluate its effectiveness, an in silico trial for asthma medications to
support randomized controlled trials, and a medical intervention for patients with Friedreich's ataxia to improve clinical decision-making and promote personalized therapy. \blfootnote{$^\dagger$ Equal contribution}\blfootnote{$^\ddagger$Work done while the author was at Princeton University}
\end{abstract}

\section{Introduction}

\epigraph{“What is spoken of the unchanging or intelligible must be certain and true; but what is spoken of the created image can only be probable; being is to becoming what truth is to belief.”}{Plato, Timaeus} 

Plato's timeless allegory of the Cave is a classical philosophical thought experiment that routinely comes up in discussions of how humans perceive reality and whether there is any higher truth to existence. A group of people live chained to the wall of a cave all their lives, facing a blank wall and watching shadows projected on the wall from objects passing in front of a fire behind them, until one one of them is freed. This prisoner imagines what would it be to look around, only to discover the real nature of the world they have perceived through the shadows thus far. The pursuit of this higher truth has enabled us with the ability to be unshackled from the perceived reality and mathematically reason about alternate, imagined perspectives, an ability coined as counterfactual reasoning. 
Such reasoning abilities form the hallmark of an agent operating in a dynamic environment and allows them to reliably manipulate the world by ascertaining the possible counterfactuals of potential interventions. Examples of such agents and interventions are ubiquitous in healthcare: policy-makers enact laws to improve public health, healthcare systems iteratively improve the quality of care they provide, and physicians determine medical treatment plans for their patients. In each of the previous cases, to make a decision, the physicians and policy-makers acting as agents need to evaluate the ability of each intervention option in the form of medications and treatment policies to yield the desired results. 

\begin{figure}[htb]
    \centering
    \includegraphics[width=0.7\linewidth]{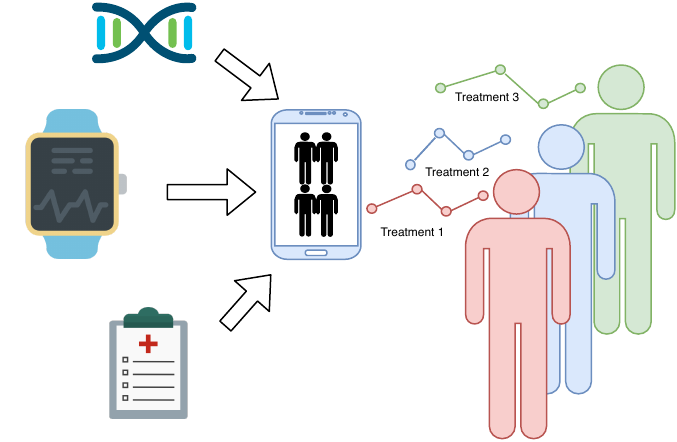}
    \caption{An individual's genetic, physiological, and clinical data can be used to synthesize their counterfactuals under multiple interventions using donor units.}
    \label{fig:precision}
\end{figure}

Often, there is interest in reliably estimating the effect of an intervention at both the population and individual levels. Our understanding of the genetic basis of diseases has accelerated in recent years and ushered in the era of precision medicine, a new paradigm in therapeutics that advocates for personalized treatments. Even patients diagnosed with the same disease may benefit from individualized therapies 
\cite{Konig1700391}. Hence, tailored clinical approaches based on insight extracted from large
amounts of data--including clinical, genetic, and physiological information--are beneficial, as depicted in 
Fig.~\ref{fig:precision}. Public health officials also use demographic and economic data to assess possible interventions and make effective decisions.


Randomized Controlled Trials (RCTs) are the current gold standard for evaluating the effect of a
treatment. In an RCT, participants are randomly assigned to either an intervention arm or a control
arm. The outcome variables of the exposed and control groups are then compared to estimate the
treatment effect. This approach is population-based and forces physicians to adopt a one-size-fits-all treatment that is prone to error. Moreover, RCTs raise a variety of issues around ethics and cost  \cite{Sibbald201}. In contrast to traditional population-based methods of evaluating treatments, precision medicine encourages personalized approaches to ``delivering the right treatments, at the right time, every time to the right person,'' a phrase famously recited by former U.S. President Barack Obama. In the absence of RCTs, social scientists rely on Comparative Case Studies (CCS) to estimate the effect of an intervention. In CCS, a group of `donor' units is collected that resemble the unit of interest prior to the onset of intervention. Then the evolution of the donors produces a counterfactual for the unit of interest.  However, finding the donors is empirically driven and the existence of a good match is not guaranteed.

\begin{figure}[htb]
    \centering
    \includegraphics[width=0.7\linewidth]{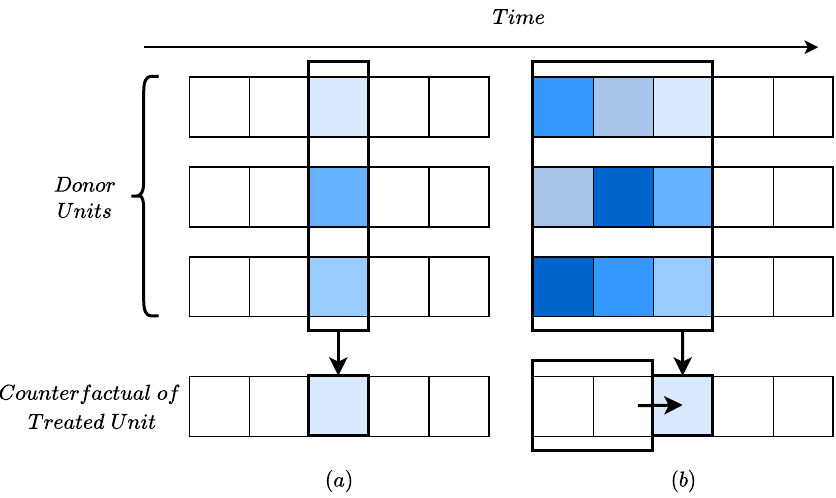}
    \caption{Synthetic control: (a) The vanilla SC method generates control via a convex interpolation of 
the donors across temporal slices using time-agnostic weights. (b) We view the problem through the lens of 
a sequence-to-sequence modeling problem and use local spatiotemporal context while predicting control. }
    \label{fig:dsc_sc}
\end{figure}

Abadie et al.~\cite{abadiebasque, doi:10.1198/jasa.2009.ap08746} pioneered a powerful
data-driven approach to aggregate observational data from several donor control units to
compute the Synthetic Control (SC) of a target unit that receives treatment. They posited
that a combination of several unaffected units provides a better match to the affected unit
than any unaffected unit alone. Hence, the SC method assigns a non-negative weight to each donor. Such weights  
sum to 1.0 such that a convex combination of these units best estimates the treated 
unit prior to the intervention. After that, the post-intervention donor combination yields
the desired counterfactual. Prior works have relaxed the convex constraints on the
vanilla SC estimator and explored regression-based estimators to compute the assigned
weights \cite{https://doi.org/10.1002/jae.1230, doudchenko2017balancing}. Moreover, recent
works \cite{JMLR:v19:17-777,amjad2019mrsc, mcnnm} use matrix estimation methods to make SC more
robust to noise and missing data. Similarly, Synthetic Intervention (SI) for a target control unit can be estimated by selecting donor units that undergo the intervention \cite{agarwalsi}. While SC can generate longitudinal trajectories of patient progress with no intervention exposure, SI can simulate trajectories under a medical intervention. The class of linear SC methods essentially solves an optimization problem for mapping
donors to the target across temporal slices using universal, time-agnostic weights. While approaches based on linearly combining donor outcomes to estimate the counterfactual yield an agile model, they overlook both the inter-unit and intra-unit
temporal context while making predictions, as illustrated in Fig.~\ref{fig:dsc_sc}. These
additional relations are essential to take into account because the evolution dynamics of a unit 
may lead to long-range temporal dependencies that can extend into other units and cause 
co-movement. We formalize this notion in Section \ref{sec:assumption}. 

In this work, we are interested in the point treatment setting where the counterfactual is estimated after an
intervention is administered once to a target unit at some time instance. We recast the problem of counterfactual prediction as a sequence-to-sequence (Seq2Seq) mapping problem where the
pre-intervention spatiotemporal context of every unit is mapped into the post-intervention sequence
of the target. From a modeling standpoint, we introduce a novel Transformer-based model
\cite{vaswani} that uses self-attention to learn complex dynamics. At its core, a Transformer is a generalized differentiable computer that is expressive in the forward pass, easily optimizable via back-propagation, and highly parallel computationally. Thus it can be used as an efficient universal compute engine wrapped around by modality-specific encoder and decoder heads to model diverse modalities simultaneously. Prior counterfactual estimation methods hinge on the provision of simple vector covariates but are unamenable to complex inputs like images and audio, whereas Transformers have shown remarkable modeling properties in these areas. Thus this choice of model opens up the avenue to do counterfactual estimation on all kinds of medical data. In Section \ref{sec:fa}, demonstrate our framework on ubiquitously occurring discrete classes, a simple abstraction that no prior works have tackled. Due to the spatial component of
the problem, we modify the decoder attention map to enable bidirectional attention spatially while
still maintaining temporal causality. Furthermore, we inject spatial and temporal embeddings to help
the model learn the input token order and add an additional target embedding to separate the target
and donor units. From a computational perspective, while SC estimates rely on a relatively small
number of donors, Transformer models traditionally require large amounts of data to be trained from
scratch. To mitigate this problem, we formulate a novel self-supervision-based pre-training task for
the Transformer model. The Transformer-based estimator captures rich nonlinear function classes, accounts for irregular
observations, and is easy to train using backpropagation. It
is, therefore, a practical non-parametric approach for modeling complex real-life data compared to linear SC estimators
whose assumption of time-invariant correlation across units 
fails for weakly coupled dynamical systems. Our experiments show that the model outperforms prior state-of-the-art estimators on 
synthetic datasets for different donor pool sizes and remains robust to noise. Running ablation studies on the pre-training step reveals that it is critical to 
learning a good estimator. We study the effectiveness of an anti-tobacco law passed in California and probe the attention layer of the Transformer model to generate a rich description of spatiotemporal 
donor weights. We also focus on the individual level and model the control counterfactual of patients in a randomized
trial on drugs for childhood Asthma. Finally, we simulate synthetic counterfactuals under a Calcitriol supplement
intervention for a Friedreich's ataxia (FA) patient. 

\section{Related Work}

Our work relates most closely to \cite{abadiebasque} which first introduced the SC method to measure the economic ramifications of terrorism on Basque Country in a panel data setting. They took observational data from other Spanish regions and assigned them convex weights to generate an interpolated SC. They also used their method to analyze the Proposition 99 Excise Tax effect on California's tobacco consumption \cite{doi:10.1198/jasa.2009.ap08746}. Since then, this method to estimate counterfactuals has gained a lot of popularity and has 
been used in the social sciences to study legalized prostitution \cite{cunningham2017}, immigration policy \cite{10.2307/43554929}, taxation \cite{10.1257/aer.103.5.1892}, organized crime \cite{https://doi.org/10.1111/ecoj.12235}, and various other policy evaluations. Other notable works that use
regression-based techniques to estimate the control using observed data from other units are the
panel data-based estimation method introduced in \cite{https://doi.org/10.1002/jae.1230} and the
additive-difference-based estimator proposed in \cite{doudchenko2017balancing}. Recently, Robust Synthetic Control (RSC) \cite{JMLR:v19:17-777} was proposed as a robust variant of the SC method. 
It sets up the problem as an unconstrained least squares regression problem with sparsity terms and therefore
extrapolates the control from the donors. In addition, the authors introduce a low-rank decomposition denoising step 
to handle missing data and noise that are ubiquitous in real-world data. Multi-dimensional Robust Synthetic Control 
(mRSC) \cite{amjad2019mrsc} extends RSC and proposes an algorithm to incorporate multiple covariates into the model. 
Ref.~\cite{mcnnm} poses the synthetic control problem as estimating missing values in a matrix. For an extensive literature review of SC, see \cite{Abadie2021UsingSC}. The authors in 
\cite{NEURIPS2021_19485224} propose a neural network-based estimator that learns a unit-specific representation and 
thereafter uses them to find donor weights. The linear SC estimators have been generalized to generate
synthetic counterfactuals under any intervention \cite{agarwalsi}.  Most of the above works make the strong
parametric assumption that the underlying data-generating process is a linear factorized model of a time-specific 
and unit-specific latent. Therefore, the proposed models combine donor outcomes across temporal slices and forgo 
any temporal context. Comparative case studies have been used historically to estimate the control 
\cite{card90,card93}. However, such works have not formalized the selection of the donor pool.

Orthogonal to SC, structural time series methods under the stringent unconfoundedness assumption use temporal regularities across control paths to extrapolate the pre-treatment trajectory to the counterfactual outcome of the treated unit \cite{rosenbaum83,imbens2009,cattaneo2018}. Structural models need many donor paths and are not amenable to settings like healthcare where the donor pool is small. Difference in Difference (DiD) methods have also been used in healthcare 
applications \cite{wing2018}, but they rely on the strong assumption that the slopes of observed post-treatment units and control units diverge only because of the treatment, such that any difference in slopes can be attributed to the intervention. Thus, non-parallel trends present a major
challenge to the validity of this model. Besides, there exists a rich literature on non-SC machine-learning-driven 
counterfactual estimation. Neural network-based techniques have been used to estimate the Individual Treatment Effect 
in both static \cite{uri2016,Yoon2018GANITEEO} and longitudinal settings \cite{bica2020}. 

While the importance of disease progression models is widely recognized, there are few works that apply
advanced modeling approaches to the task. Many methods use basic statistics and rely on a large amount of
domain knowledge. More sophisticated probabilistic or statistical models exist but tend to be exclusively
applicable to a specific target disease. Using a more general framework, ref.~\cite{wang2014} predicts the
stage of chronic obstructive pulmonary disease progression with a graphical model based on the Markov Jump
Process. One Transformer
model under the broad umbrella of disease progression models has been proposed;
\cite{nguyen2021} predicts the development of knee osteoarthritis with a Transformer model designed for multi-agent 
decision-making but relies on imaging data. In addition, most neural network approaches to disease progression only predict one to a few metrics of disease, highlighting an area that requires further research.

Many works use neural networks as Seq2Seq models for machine translation using feed-forward networks \cite{bengio2003}, recurrent neural networks \cite{Mikolov2010RecurrentNN}, and long short-term memories \cite{sutskever2014sequence}. Transformer-based models have achieved state-of-the-art results on tasks that span natural language processing (NLP) \cite{vaswani} and, more recently, computer vision \cite{vit_2021}. They are pre-trained on large corpora and then fine-tuned for a specific task, with the most notable works being BERT \cite{bert}, which uses self-supervision-based denoising tasks, and GPT \cite{Radford2019LanguageMA,brown2020language}, which uses a language-modeling-based pre-training task. Our work adds to these works by proposing a model to map spatiotemporal data and introducing a self-supervision-based pre-training task that supports the mapping.

\section{Background}

In this section, we  outline the key assumptions and formally set up the problem of generating synthetic counterfactuals using donor units. 

\subsection{Assumptions}
\label{sec:assumption}
The problem consists of a `target' unit and $\mathit{N}$ `donor' units with $\mathit{N}$ typically small. A unit can 
refer to an aggregate of a population segment or an individual. $\mathit{N}$ donor units undergo the intervention we 
are interested in at the same time instance. 
Without loss of generality, we assume that control assignment is also a type of intervention and finding the 
counterfactual under control reduces to the SC problem. For each unit, panel data are available for a time 
period $\mathit{T}\;$ and the covariate in each panel can have missing data. We make standard assumptions about the units. 
\begin{enumerate}
    \item \textbf{Stable Unit Treatment Value Assumption} (SUTVA): The potential outcomes for any unit do not vary with the treatments assigned to other units and, therefore, there are no spillover effects.
    \item \textbf{No anticipation of treatment:}  This implies that 
the units cannot anticipate the assignment of treatment {\em a priori} and, therefore, the assignment has no effect on the pre-intervention trajectory.
    \item \textbf{Exchangeability} (No unmeasured confounders): This assumption implies that units that are exposed and unexposed have the same potential outcomes on average and hence in the case of an observational study, a control group can be reliably used to measure the counterfactual of the treatment group.
    \item \textbf{Positivity} (Common support): This necessitates covariates in the control and the treatment group have overlapping distributions (common support), in the absence of which it is impossible to understand the causal effect of a treatment because a subset of the population is always entirely left treated or untreated.
\end{enumerate}  

\textbf{Data-generation Process:} The underlying data generation of a unit $i$  consists of unit-specific latent $\theta_i$ 
and time-indexed latents $\rho_t$. The observation of all covariates of unit $i$ at time $t$, 
$y_{it}$, depends on all prior temporal latents and is given by:

\begin{gather}
    y_{it} = m_{it} + \epsilon_{it}, \epsilon_{it} \sim \mathcal{N}(0,\sigma^2).\\
      m_{it} = f(\theta_i,\rho_t , \cdots, \rho_{1}).
      \label{eq:dgp}
\end{gather}
       
Function $f$ is implicitly learned by the Transformer that can capture a wide class of nonlinear autoregressive dynamics. This generalized model that subsumes the models popularly assumed in the literature. Ref. \cite{abadiebasque} assumes 
the following model for the data-generation process.  
  \begin{equation}
      y_{it} = {\rho_t}^TX_i + \epsilon_{it}. \label{eq:scdgp}
  \end{equation}
Here, $\rho_t$ denotes the latent at time $t$ and $X_i$ denotes the observed covariates of unit $i$. Setting
$f$ to the linear factor function and $\theta_i = X_i$ in Eq.~(\ref{eq:dgp}) yields the model defined by 
Eq.~(\ref{eq:scdgp}). Similarly, ref.~\cite{NEURIPS2021_19485224} assumes that:
  \begin{equation}
      y_{it} = {\rho_t}^TC_i + \epsilon_{it}. \label{eq:scdgp}
  \end{equation}
  
\noindent
where $C_i$ is a latent unit-specific representation. It is straightforward to see that Eq.~(\ref{eq:dgp}) captures 
this model as well. Refs.~\cite{JMLR:v19:17-777,amjad2019mrsc} assume a non-autoregressive, nonlinear $f$ that is 
once again a special case of our assumption. Next, we set up the problem formally.

\subsection{Problem Setup}
The synthetic counterfactual problem uses a `target' unit and $\mathit{N}$ `donor' units. For the SC, the target
unit $\mathit{Y}$ is exposed to a treatment or intervention, and the donor units are assumed to be unexposed
and, therefore, natural controls. Therefore, we are interested in the counterfactual of the target under the
control assignment. On the contrary, donors are exposed to the intervention at a time instance and are natural 
controls prior to that instance for computing the SI. Therefore, SC and SI are dual problems and simply differ in 
the type of intervention the donors receive at the intervention instance. 

Each unit is $\mathit{T}$ time units long,
and at each time step, the sequence has $\mathit{K}$ covariates, e.g., Gross Domestic Product,
literacy rate, or physiological data, where each covariate may have missing values. Let each element in the target sequence $Y \in \mathbb{R}^{T \times K}$ be denoted by $y^{t,k}$, where $\mathit{t}$ denotes the time instance and $k$ denotes the covariate. Without loss of generality, assume that the covariate of interest is at $k=1$, and let $\mathit{T_0}$ be the time of intervention. The target sequence $Y=[Y^-,Y^+]$ is thus divided into a pre-intervention sequence $\mathit{Y^-} \in \mathbb{R}^{T_0 \times K}$ and a
post-intervention sequence $\mathit{Y^+}  \in \mathbb{R}^{(T-T_0) \times K}$. Let the tensor $X \in \mathbb{R}^{N \times T \times K}$ represent the data from $\mathit{N}$ donor units and $x^{i,t,k}$ represent each element of the donor tensor $X$, where $i$ is the donor, $t$ is the time instance, and $k$ is the covariate. Let $X=[X^-,X^+]$, where $X^- \in \mathbb{R}^{N \times T_0 \times K} $ and $X^+ \in \mathbb{R}^{N \times (T-T_0) \times K} $ denote the pre-intervention and post-intervention data of the donors, respectively. The SC problem involves learning a predictor $f^t_\theta(.)$ for the post-intervention control of the covariate of interest, $\widehat{y}^{t,1} \; \forall t \in [T_0+1, \cdots ,T]$, using the  donor data $X$ and the pre-intervention target sequence $Y^{-}$. More formally:
\begin{equation}
    \widehat{y}^{t,1} = f^t_{\theta}\left(X,Y^{-}\right), \: \forall t \in [T_0+1, \cdots ,T].
\end{equation}

\subsection{Transformers}
Transformers \cite{vaswani} were proposed to model sequential data. They have achieved state-of-the-art 
results on several NLP tasks. The model consists of stacked layers, where each layer has a self-attention 
module followed by a feedforward module, and each module has residual connections \cite{He_2016_CVPR} 
and layer normalization \cite{ba2016layer}. The self-attention module learns powerful internal 
representations that capture important semantic and syntactic associations across input tokens 
\cite{Manning30046}. This module decomposes each token into a query ($Q$), key ($K$), and value ($V$) 
vector, and uses these vectors to aggregate global information at each sequence position. For finer 
details of this architecture, we refer readers to the original article.

\subsection{Benchmarks}
\label{sec:benchmark}
In this section, we describe prior methods to estimate counterfactuals for longitudinal outcomes under a point
treatment setting. Since the introduction of the SC technique \cite{abadiebasque}, several modifications to it
have been proposed. We describe several state-of-the-art estimators that we benchmark our method against. 



 
\subsubsection{Robust Synthetic Control (RSC) }
\label{sec:RSC} The RSC method \cite{JMLR:v19:17-777} assumes a linear model and introduces a denoising step to the SC method. Furthermore, it uses regularized regression to find the optimal weights $\beta^*$. Let $\hat{p}$ be the fraction of missing data in $X$. The authors replace missing data in $X$ with $0$ and estimate a low-rank matrix $\hat{M}$ from $X$ as follows:
\begin{gather}
    X = \sum s_i u_i v_i^T. \\
    S = \{ i: s_i \geq \mu \}. \\
    \hat{M} = \frac{\sum_{i \in S} s_i u_i v_i^T }{\max(1-\hat{p},\frac{1}{NT})}.
\end{gather}

The threshold $\mu$ above is a hyperparameter. Let $\hat{M} = [\hat{M}^-,\hat{M}^+]$, where $\hat{M}^-$ 
and $\hat{M}^+$ denote the pre-intervention and post-intervention data, respectively. The optimal weights 
$\beta^*$ are computed using the LASSO regularized regression as follows:
\begin{gather}
    \beta^* = \argmin_{\beta} \left\lVert Y^- - M^{-T} \beta \right\rVert ^2 + \eta \left\lVert \beta \right \rVert_{1} .
\end{gather}
The counterfactual is predicted through
\begin{gather}
    \widehat{Y} = \hat{M}^{+T}\beta^*.
\end{gather}
 
 \subsubsection{Multi-Dimensional Robust Synthetic Control (mRSC)}
 
 \label{sec:mrsc}
The same authors \cite{amjad2019mrsc} generalize RSC to incorporate multiple additional covariates to predict the covariate of interest. At a high level, they flatten their donor tensor $X \in \mathbb{R}^{N\times T \times K}$ to a matrix $Z \in \mathbb{R}^{N \times KT}$. After that, they run the denoising step from RSC on $Z$ and perform a covariate-based re-weighting on the resultant low-rank matrix and target matrix $Y$. In their final step, they run regression on the pre-intervention data of the re-weighted matrices. 

\subsubsection{Matrix Completion using Nuclear Norm Minimization (MC-NNM)}

The authors in \cite{mcnnm} estimate the counterfactual by treating them as missing values of a matrix. Let $Y \in \mathbb{R}^{T \times (N+1)}$ denote the outcome matrix of the donor and target units. Then the underlying structure assumed on the matrix $Y$ by the authors is:
\begin{gather}
    Y = L^* + \epsilon\\
    L^* = \hat{L} + \hat{\tau}\mathbf{1}_T^T + \mathbf{1}_N\hat{\Delta}^T
\end{gather}

Here $L^*$ denotes the true mean matrix, $\hat{\tau}$ models the fixed temporal effect, $\hat{\Delta}$ accounts for the unit wide fixed effect and $\mathbf{1}$ indicates a vector of ones.
Thereafter, the true matrix $L^*$ is recovered by setting up the following optimization problem with a penalty on the nuclear norm.

\begin{gather}
    (\hat{L},\hat{\tau},\hat{\Delta}) = \argmin_{L,\tau,\Delta} \lVert Y - L -\tau \mathbf{1}_T^T - \mathbf{1}_N\Delta^T \rVert_{F} + \lambda \lVert L\rVert
\end{gather}

\subsubsection{Sync-Twin}
The authors in \cite{NEURIPS2021_19485224} propose a learning algorithm to learn a deep representation $c_i$ for the units using pre-intervention covariates. They assume the following factorised data-generation model:

\begin{equation}
    Y_{it} = q_t^Tc_i + \epsilon_{it}
\end{equation}

\noindent
where $q_t$ denotes the temporal factors. Thereafter, $c_i$ is learned by setting up the following reconstruction task over the covariates    

\begin{equation}
    \hat{c_i} = arg min_{c_i}\lVert (\tilde{X}_i - X_i)\odot M \rVert 
\end{equation}

Here, $M$ is a mask for missing covariates. Since the underlying model is linear in $\hat{c}_i$, the donor weights are calculated by regressing in the representation space $C$:

\begin{equation}
      \beta^* = \argmin_{\beta} \left\lVert c_{treated} - \hat{C}^{T} \beta \right\rVert ^2 
\end{equation}

\section{Spatiotemporal Seq2Seq Modeling}
In our work, we view the problem through the lens of Seq2Seq modeling. For the intervention instance $T_0$, 
we denote the counterfactual of the target until time $t$ as $\hat{Y}^t = 
[\hat{y}^{T_0},\hat{y}^{T_0+1}, \cdots , \hat{y}^{t}]$ and similarly the post-intervention of the donor data until $t$ as $X^{+t}\in \mathbb{R}^{N \times (t-T_0) \times K}$. Then we model the distribution of the synthetic counterfactual as:
\begin{gather}
    P(\hat{Y}|X,Y^-) = \prod_{t=T_0+1}^T P_{\theta}(\hat{y}^t|\hat{Y}^{t-1},X^{+t},X^-,Y^-).
\end{gather}
We use an encoder-decoder model, where the encoder $f_\alpha(.)$ computes a hidden representation $\mathcal{V}$ for the pre-intervention data $\langle Y^-,X^- \rangle$ and passes it to the decoder $g_\phi(.)$  that uses the representation and the post-intervention donor data $X^{+t}$ to autoregressively generate the control $\hat{y}^t$.
\begin{gather}
    \mathcal{V}= f_\alpha(X^-,Y^-).\\
   P_{\theta}(\hat{y}^t|\hat{Y}^{t-1},X^{+t},X^-,Y^-) = g_\phi(\hat{Y}^{t-1},X^{+t},\mathcal{V}).
\end{gather}
\subsection{Architecture}
\label{sec:architecture}
\begin{figure*}[thb]
    \centering
    \includegraphics[width=\textwidth]{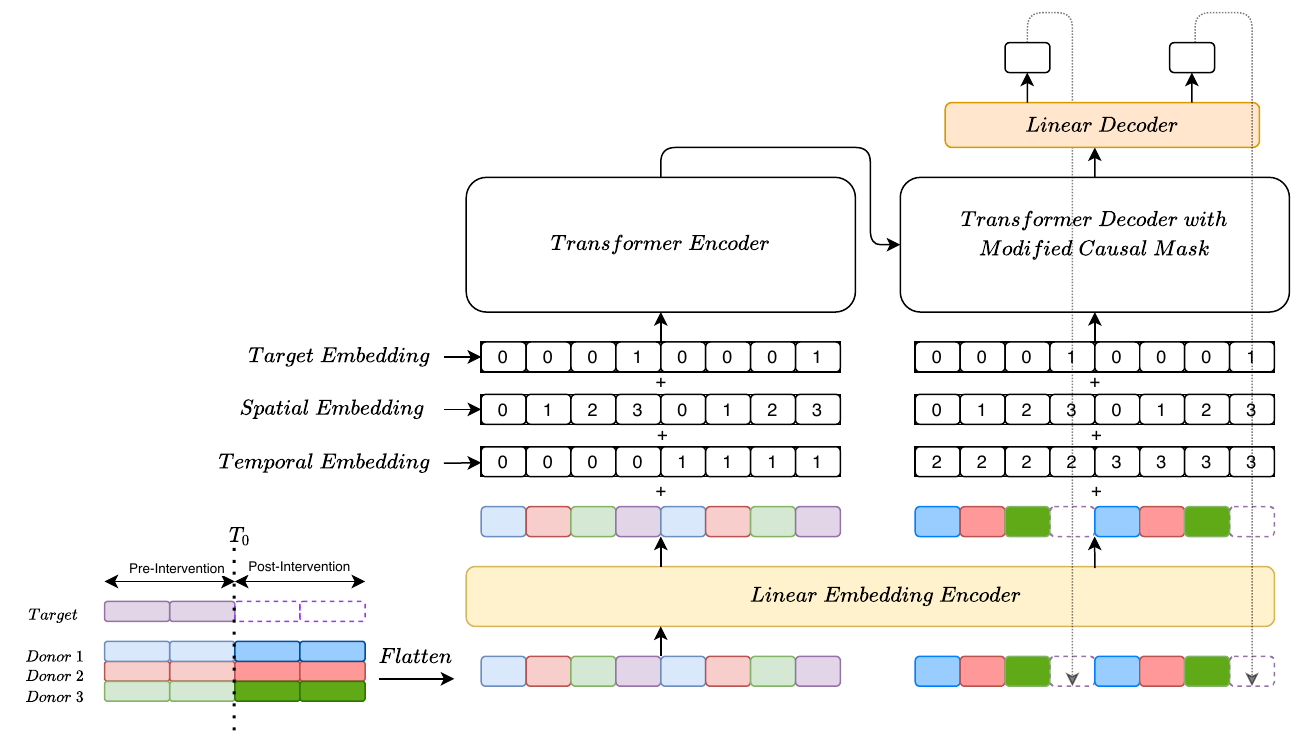}
    \caption{An overview of our model. We flatten the pre-intervention and post-intervention data into sequences and linearly embed them. Next, we add positional information in the form of temporal and spatial embeddings, and help the model differentiate between the target and donor units by injecting a target embedding. The resultant sequence of vectors is fed to the Transformer-based encoder-decoder model. The decoder uses a modified causal map, which enables spatial bidirectionality, to autoregressively generate SC of the treated unit.}
    \label{fig:architecture}
\end{figure*}

An overview of our Transformer model is given in Fig.~\ref{fig:architecture}. It encodes pre-intervention data of temporal length $l^-$ and decodes it into post-intervention data of temporal context $l^+$. The pre-intervention data of the target and the donors can be represented by the tensor $Z^-=[Y^-;X^-] \in \mathbb{R}^{(N+1) \times l^- \times K}$. Similarly, the post-intervention data are represented by $Z^+=[\hat{Y};X^+] \in \mathbb{R}^{(N+1) \times l^+ \times K}$. The standard Transformer receives a 1D sequence of input tokens. Hence, we flatten $Z^-$ and $Z^+$ into a 1D sequence $Z_{flat}^- \in \mathbb{R}^{(N+1)l^- \times K}$ and $Z_{flat}^+ \in \mathbb{R}^{(N+1)l^+ \times K}$, respectively. We then project each token into the hidden dimension $D$ of the Transformer via a trainable linear weight $W_e \in \mathbb{R}^{K \times D}$ to give sequences $E^-\in \mathbb{R}^{(N+1)l^- \times D}$ and $E^+\in \mathbb{R}^{(N+1)l^+ \times D}$. More precisely, 
\begin{align}
E^- &=  Z^-_{flat}W_e = \left[x^{i,t}W_e; \cdots x^{N,t}W_e; y^{-t}W_e\right]_{t=T_0- l^-+1}^{T_0}.  \\
E^+ &= Z^+_{flat}W_e=\left[x^{i,t}W_e; \cdots x^{N,t}W_e; \hat{y}^{t}W_e\right]_{t=T_0+1}^{T_0+l^+}.
\end{align}

\subsubsection{Positional and Target Embeddings}
Each token in the sequence belongs to one of $N+1$ tokens, one of $T$ time instances, and either a donor or the
target unit. We, therefore, inject a learnable spatial embedding $\mathbb{E}_{spatial}(.) \in \mathbb{R}^{(N+1)
\times D}$, time embedding $\mathbb{E}_{t}(.) \in  \mathbb{R}^{T \times D}$, and a target embedding
$\mathbb{E}_{target}(.) \in  \mathbb{R}^{2 \times D}$ to enable the model to differentiate between
spatiotemporal positions of the token in the data matrix as well as separate donors from the target unit. The
resultant encoder input $\mathit{H^-}$ and the decoder input $\mathit{H^+}$ sequences are obtained as follows:
\begin{gather}
    H^- = E^- + \mathbb{E}_{spatial}(E^-)+\mathbb{E}_t(E^-)+\mathbb{E}_{target}(E^-).\\
  H^+ = E^+ +\mathbb{E}_{spatial}(E^+)+\mathbb{E}_t(E^+)+\mathbb{E}_{target}(E^+).
\end{gather}

\subsubsection{Encoder}
We use the vanilla bidirectional Transformer encoder with a latent dimension $\mathit{D}$ consisting of $l$ stacked identical layers. The encoder processes the input sequence $H^-$ and outputs a sequence of representations $\mathcal{V}$ over the input. A key $K$ and value $V$ vector are computed over each of the tokens in $\mathcal{V}$ and passed to the decoder.

\subsubsection{Decoder}
\begin{figure}[!htb]
    \centering
    \includegraphics[width=0.8\linewidth]{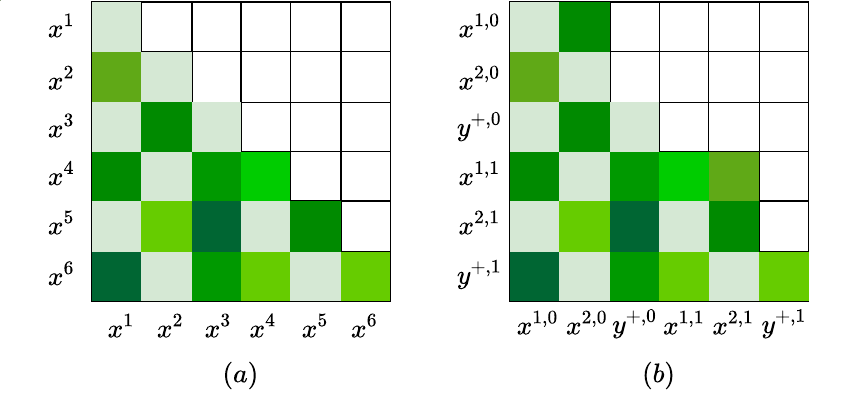}
    \caption{Comparison of the Vanilla decoder mask: (a) that enforces temporal causality and
modified causal mask, and (b) that additionally allows spatial bidirectionality.}
    \label{fig:attention_map}
\end{figure}

The decoder is tasked with autoregressively generating the counterfactual $\hat{Y}$, given the encoder
output $\mathcal{V}$ and sequence $H^+$. Our decoder design mostly follows the vanilla Transformer
decoder with $l$ stacked identical layers. Each layer has two kinds of attention, viz.,~causal 
self-attention module that operates over the decoder hidden states and the `encoder-decoder' attention module that operates over the joint representation of the encoder and decoder. However, we do modify the causal attention mask used in the self-attention module to account for the tokens that lie on the same temporal slice but differ spatially. In other words, we enforce temporal causality but allow bidirectionality spatially. The modified causal mask for spatiotemporal data is illustrated in Fig.~\ref{fig:attention_map}. We then project the hidden state of SC via a linear weight $W_d \in \mathbb{R}^{D \times 1}$ to obtain the prediction. 

Algorithm \ref{alg:transformers} lays out the skeleton of our model.

\begin{algorithm}[!htb]
\caption{Transformer $(X^-,X^+,Y^-,Y^+)\rightarrow \hat{Y}$}
\label{alg:transformers}
\begin{algorithmic}
\STATE {\bfseries Require:} $Encoder_\alpha$, $Decoder_\phi$, $W_e$, $W_d$,
$\mathbb{E}_{spatial}$, $\mathbb{E}_{t}$, $\mathbb{E}_{target}$
\STATE $Z^-_{flat} =    flatten\left(concat\left(X^-,Y^-\right)\right)$  \COMMENT{$(N+1)l^- \times K$}
\STATE $Z^+_{flat} =flatten\left(concat\left(X^+,Y^+\right)\right)$\COMMENT{$(N+1)l^+ \times K$}
\STATE $E^- =  Z^-_{flat}W_e$ \COMMENT{$(N+1)l^- \times D$}
\STATE $E^+ =  Z^+_{flat}W_e$ \COMMENT{$(N+1)l^+ \times D$}
\STATE  $H^- = E^- + \mathbb{E}_{spatial}(E^-)+\mathbb{E}_t(E^-)+\mathbb{E}_{target}(E^-)$
\STATE  $H^+ = E^+ +\mathbb{E}_{spatial}(E^+)+\mathbb{E}_t(E^+)+\mathbb{E}_{target}(E^+)$
\STATE $\mathcal{V} = Encoder_\alpha(H^-)$
\STATE $Z = Decoder_\phi(H^+, \mathcal{V})$
\STATE $\hat{Z} = Z.target$ \COMMENT{Target Hidden States}
\STATE $\hat{Y} = \hat{Z}W_d$ \COMMENT{$l^+ \times K$}
\STATE {\bfseries return $\hat{Y}$}

\end{algorithmic}
\end{algorithm}


\subsection{Data Preprocessing} 
\label{sec:preprocess}
We replace missing data with $0$ and scale each covariate in the data between $0$ and $1$ before processing by the Transformer. Motivated by mRSC \cite{amjad2019mrsc}, we apply a low-rank transformation to the data tensor $X \in \mathbb{R}^{N \times T \times K}$ when data are missing. We flatten $X$ to $X_{flat} \in \mathbb{R}^{ N \times TK}$ and retain the top $m$ singular values of $X_{flat}$.





\subsection{Pseudo-Counterfactual Prediction Pre-training}
Transformer-based language models are usually pre-trained on an unsupervised task
\cite{bert,Radford2019LanguageMA} to learn high-capacity representations that help boost
downstream performance on discrimination tasks. In the same way, we pre-train the model on
donor data to reliably reconstruct the counterfactual, the ground truth of which is known.
In each training iteration, we sample a donor unit $X_i \in \mathbb{R}^{1 \times T \times
K}, i \in [1, \cdots , N]$ and an intervention time $T' \in [1, \cdots , T]$. We treat the
sampled donor as a pseudo-target and task our model with generating its post-intervention
counterfactual. Let $X_{\backslash i}^- \in \mathbb{R}^{(N-1) \times l^- \times K}$ and
$X_{\backslash i}^+ \in \mathbb{R}^{(N-1) \times l^+ \times K}$ denote the pre-intervention
and post-intervention donor data excluding the sampled donor, respectively, and let
$\hat{X}_i^t \in \mathbb{R}^{(N-1) \times (t-T') \times K}$ denote the post-intervention control of the pseudo-target until time $t$ and $\hat{x}_i^t$ be the control at instance $t$. The objective of the model is to maximize the following log-likelihood:

\begin{equation}
    \mathcal{L}(\alpha,\phi) = \sum_{t=T'+1}^{T'+l^+} log ( P (\hat{x}_i^t| \hat{X}_i^{t-1}, X_{\backslash i}^{+t}, X^-, \alpha, \phi ))
\end{equation}

We use the teacher forcing algorithm \cite{teacherforcing} while training and assume a Gaussian model for the likelihood that reduces the loss function to squared error. 

\subsection{Fine-tuning}
Fine-tuning proceeds by fitting the model on the pre-intervention data $Y^-$ of the target unit using the pre-intervention donor data $X^-$. In each fine-tuning iteration, we sample a time instance $T'$ in 
$[1, \cdots , T_0]$ for use as the pseudo-intervention instance. Then, we treat $X^-$ and $Y^-$ as the data in hand and divide it into pseudo-pre-intervention and pseudo-post-intervention data to predict the pseudo-post-intervention of the target.

\subsection{Inference}
We generate the SC $\hat{Y}$ in a sliding window fashion, where we start at $T_0$ and generate
post-intervention data of temporal length $l^+$ each time. The generated control is used as
pre-intervention data for the subsequent time steps. Algorithm \ref{alg:trainpredict} shows the
pseudo-code for training and inference.

\begin{algorithm}
\caption{Training and Inference}
\label{alg:trainpredict}
\begin{algorithmic}
\STATE {\bfseries Require:} Transformer($\alpha,\phi$), Learning rate $\lambda$
\STATE $\triangleright$ {\bfseries Pre-training}($X$, $iters$)
\FOR{$j=1$ {\bfseries to} $iters$}
\STATE $i = random.sample(N)$
\STATE $T_0 = random.sample(T)$
\STATE $Y\;=\;X[i]$ \COMMENT{Pseudo Target}
 \STATE $X'\;=\;X.remove(i)$\COMMENT{Pseudo Donor}
 \STATE  $Y^-,\; Y^+=\;Y\;[T_0-l^-:T_0],\; Y\;[T_0:T_0+l^+]$ 
\STATE $X^-,\;X^+ = X'[:,T_0-l^-:T_0],\;X'[:,T_0:T_0+l^+] $
 \STATE $preds$ = Transformer($X^-,X^+,Y^-,Y^+$)
 \STATE $\mathcal{L}$ = $mean \left((Y^+ -preds)^2\right)$
 \STATE  $\alpha \leftarrow \alpha - \lambda \nabla_{\alpha}\mathcal{L},\;\phi \leftarrow \phi - \lambda \nabla_{\phi}\mathcal{L}$ 
\ENDFOR
\STATE $\triangleright$ {\bfseries Fine-tuning}($X,Y,T_0$, iters)
\FOR{$j=1$ {\bfseries to} $iters$}
\STATE $T' = random.sample(T_0)$
\STATE $Y'\;=\;Y\;[ :T_0]$ 
 \STATE $X'\;=\;X\;[:,:T_0]$
 \STATE  $Y^-,\;Y^+=\;Y'\;[T'-l^-:T'],\;Y'\;[T':T'+l^+]$ 
\STATE $X^-,\;X^+ = X'[:,T'-l^-:T'],\;X'[:,T':T'+l^+] $ 
 \STATE $preds$ = Transformer($X^-,X^+,Y^-,Y^+$)
 \STATE $\mathcal{L}$ = $mean \left((Y^+ -preds)^2\right)$
 \STATE  $\alpha \leftarrow \alpha - \lambda \nabla_{\alpha}\mathcal{L},\;\phi \leftarrow \phi - \lambda \nabla_{\phi}\mathcal{L}$ 
\ENDFOR
\STATE $\triangleright$ {\bfseries Inference}($X,Y^-,T_0$) $\rightarrow \hat{Y}$
 \STATE $X^-\;=\;X\;[:,T_0-L^-:T_0]$
\STATE  $\hat{Y}$ = [ ]
    \FOR{$I=1$ {\bfseries to} $l^+$}
     \STATE $X^+\;=\;X\;[:,T_0:T_0+i]$ 
     \STATE $preds$ = Transformer($X^-,X^+,Y^-,\hat{Y}$)
     \STATE  $\hat{Y} = \hat{Y}.append(preds[i])$
    \ENDFOR
\STATE {\bfseries return $\hat{Y}$}
\end{algorithmic}
\end{algorithm}

\subsection{Visualization}
\label{sec:interpretability}

A key property of the simple linear model assumed by the vanilla SC method is its interpretability in the form of donor weights. Modeling the spatiotemporal context helps us visualize the donor contributions across different time instances. Transformers are effective at long range credit assignment and the attention score $a_{ij}$ in the attention matrix $A$ of the Transformer specifies the proportion with which a token $i$ attends to token $j$. We probe the self-attention layers of the decoder and extract the attention scores of various donors while predicting the target token. While not entirely dispositive of the donor significance, this can help domain experts uncover the relations between targets and important donors.

We qualitatively compare our proposed approach against prior techniques in Table \ref{tab:qual} with respect to four 
features. 1) \textbf{Nonlinearity:} Whether the model captures nonlinear dynamics.  2) \textbf{Missing Data:}
Whether the proposed algorithm considers missing data values for the outcome variable and covariates. 3)
\textbf{Temporal Context:} Whether the model implicitly or explicitly considers the prior context. 
4) \textbf{Multimodal Inputs:} Whether multimodal data like images, audio, and discrete classes can be processed 
by the proposed approach. 

\begin{table}[htb]
\caption{Qualitative comparison of past approaches and our method. \cmark \; indicates the presence of a feature and \xmark \; denotes its absence. *Sync Twin only considers missing pre-intervention covariates.}
\label{tab:qual}
\vskip 0.15in
\begin{center}
\begin{tabular}{ccccc}
\toprule
Method  & \thead{Non\\Linearity} & \thead{Missing\\Data} & \thead{Temporal \\Context} & \thead{Multimodal \\ Inputs}\\
\midrule
\thead{Synthetic Control \cite{abadiebasque}}  & \xmark & \xmark & \xmark & \xmark  \\
RSC \cite{JMLR:v19:17-777} & \xmark & \cmark & \xmark & \xmark\\
mRSC \cite{amjad2019mrsc} & \xmark & \cmark & \xmark & \xmark\\
MC-NNM \cite{mcnnm} & \xmark & \cmark & \cmark & \xmark\\
Sync-Twin \cite{NEURIPS2021_19485224} & \cmark & \xmark\textsuperscript{*} & \xmark & \xmark\\
\midrule
\thead{Spatiotemporal\\Transformer \\(Ours)} & \cmark & \cmark & \cmark & \cmark\\

\bottomrule
\end{tabular}
\end{center}
\end{table}
\section{Experiments}
We demonstrate the efficacy of our method through quantitative and qualitative experiments. We give the training 
details, additional results and hyperparameters for each experiment in the Appendix.
\subsection{Synthetic Data}

Counterfactual estimators are difficult to evaluate on real-life data since the ground truth counterfactual is
unknown. Hence, we benchmark our method on synthetic data where the underlying data-generation process is known 
{\em a priori}.  We set up an SC problem here by constraining the donors to be controls.
The `mean' synthetic data $M$ is generated using a latent factor model. $M \in \mathbb{R}^{(N+1) \times T 
\times 2}$ has two covariates and the first row represents the target data with the remaining tensor 
representing donor data. Each row $i$ and column $j$ is assigned a latent factor $\theta_i$ and 
$\rho_j$, respectively, and covariate $k$ is generated using $f_k(\theta_i,\rho_j)$. We use

\begin{gather}
    \begin{split}
    f_1(\theta_i,\rho_j) =  \theta_i + \frac{(\rho_j \times \theta_i)}{T}exp(\frac{\rho_j}{T})+cos(2\mathcal{F}_1\frac{\pi}{180}) + \\sin(\mathcal{F}_1 \frac{\pi}{180}) + cos(2\mathcal{F}_2\frac{\pi}{180})+sin(\mathcal{F}_2\frac{\pi}{180}).
    \end{split}
    \\
    f_2(\theta_i, \rho_j) = \frac{10}{1+exp(-\theta_i - \frac{\rho_j}{T}-0.7 \frac{\theta_i \rho_j}{T})}.
\end{gather}

Here, $\mathcal{F}_1 = (\rho_j\;mod\;360),\; \mathcal{F}_2 = (\rho_j\;mod\;180)$. Thereafter, we add noise 
$\epsilon \sim N(0,\sigma^2)$ to each entry in $M$ to obtain the observed matrix $O$. Given an intervention 
instance $T_0$, we task the estimator to reconstruct the post-intervention ($t > T_0$) mean sequence of 
the first covariate from noisy observations. Note that our data-generation process is similar to 
\cite{amjad2019mrsc} with the critical distinction being that the mean target sequences are generated from 
a high-rank latent factor model instead of being linearly extrapolated from donor units composed from a 
low-rank latent factor model. This makes the task inherently `harder'. For all the experiments run on 
synthetic data, we set $T=2000,\; T_0 = 1600,\;\rho_j =j$, and $\theta_i$ is uniformly sampled from 
$[0,1]$. Our proposed method is compared against all the benchmark methods mentioned in Section \ref{sec:benchmark}, using the root mean squared error (RMSE) between the estimates and the true post-intervention mean. In addition, to this we also learn a simple Transformer based decoder baseline to perform time series modeling on the donor trajectories. This baseline therefore corresponds to the extreme case of zero donors used to extrapolate the counterfactual under the modeling framework.

\subsubsection{Robustness against Noise}
\begin{figure}[!htb]
    \centering
    \includegraphics[width=\linewidth]{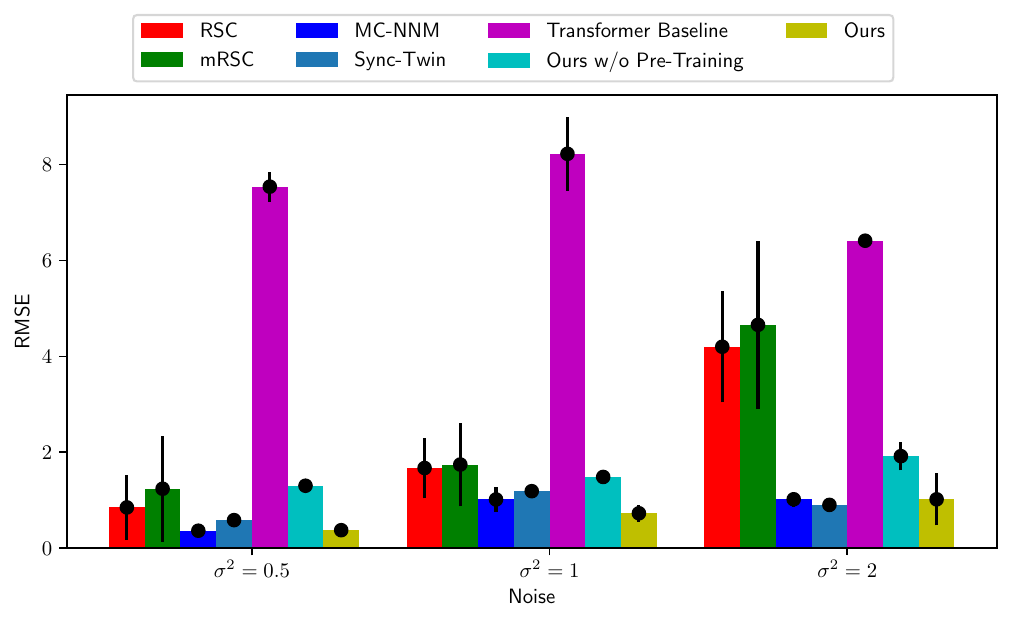}
    \caption{Comparison between post-intervention RMSE of our approach and prior estimators 
using observations with varying noise levels. Our approach surpasses or performs at par with these techniques at all levels of noise. Moreover using donors and therefore spatiotemporal transformers (yellow) gives $\mathbf{6x - 20x}$ reduction in error over simple temporal modeling of control path using Transformers (purple). Error bars are plotted at 90\% confidence intervals.}
    \label{fig:noise}
\end{figure}

\begin{figure*}[!htb]
    \centering
    \includegraphics[width=\textwidth]{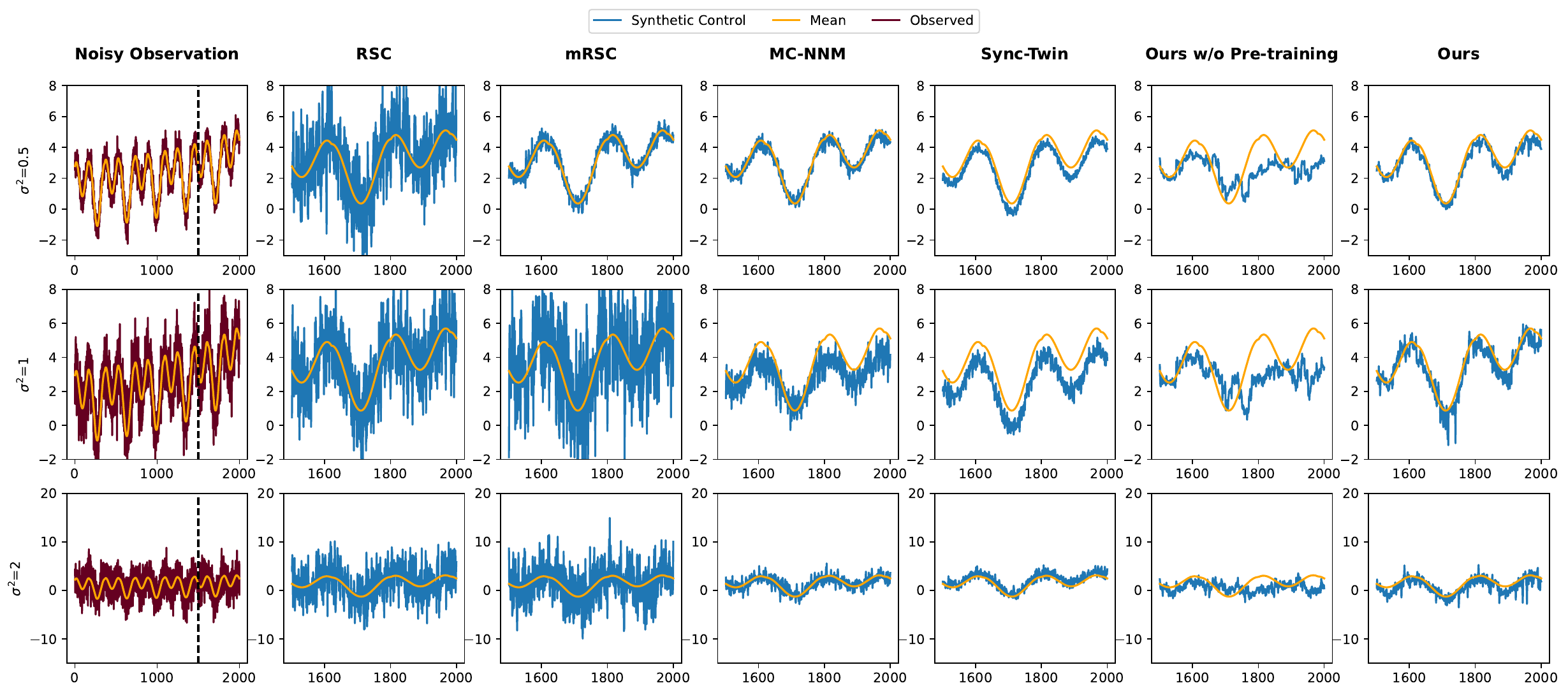}
    \caption{Estimates of the post-intervention mean of the target from noisy observations for $N=10$ and varying levels of noise. The vertical black line indicates the intervention instance.}
    \label{fig:noise_all}
      \vspace{-1em}
\end{figure*}

Next, we investigate the robustness of our method against noise in the observed data. We set $N=10$
and generate three synthetic datasets with varying noise levels $\sigma^2 \in [0.5,\;1,\;2]$. The
RMSE scores are reported in Fig.~\ref{fig:noise} and Fig.~\ref{fig:noise_all} shows the synthetic control estimates. The Transformer baseline performs the worst, which is expected given traditional transformers' data inefficiency. This also indicates the difficulty of using simple time series extrapolation with few donors and shows the benefits of directly modeling the donor outcomes to perform few-shot counterfactual estimation. As noise increases, our predictions remain robust and
lie close to the true mean without using any denoising step.  Moreover, pre-training is essential for 
helping our model learn a useful representation of the underlying latent model, thus in making 
accurate predictions.  

\subsubsection{Effect of the donor pool size}
\label{sec:small_donor}

\begin{figure}[!htb]
    \centering
    \includegraphics[width=\linewidth]{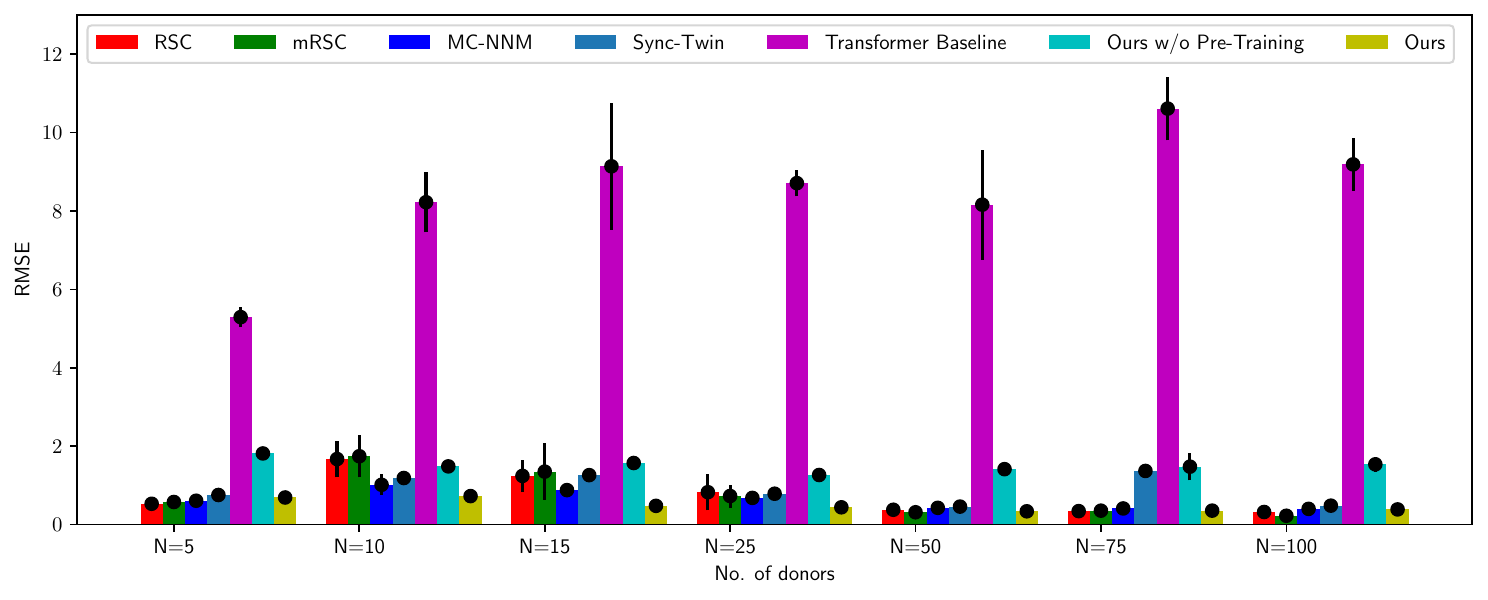}
    \caption{Post-intervention RMSE of various estimators with varying donor pool size. Our approach performs competitively for most donor pool sizes with the pre-training step being crucial to achieving an accurate prediction. Using spatiotemporal transformers (yellow) reduces error over a traditional Transformer (purple) by $\mathbf{6x-19x}$. Plotted at 90\% confidence intervals. }
    \vspace{-1em}
    \label{fig:donorpool}
\end{figure}

\begin{figure*}[!htb]
    \centering
    \includegraphics[width=\textwidth]{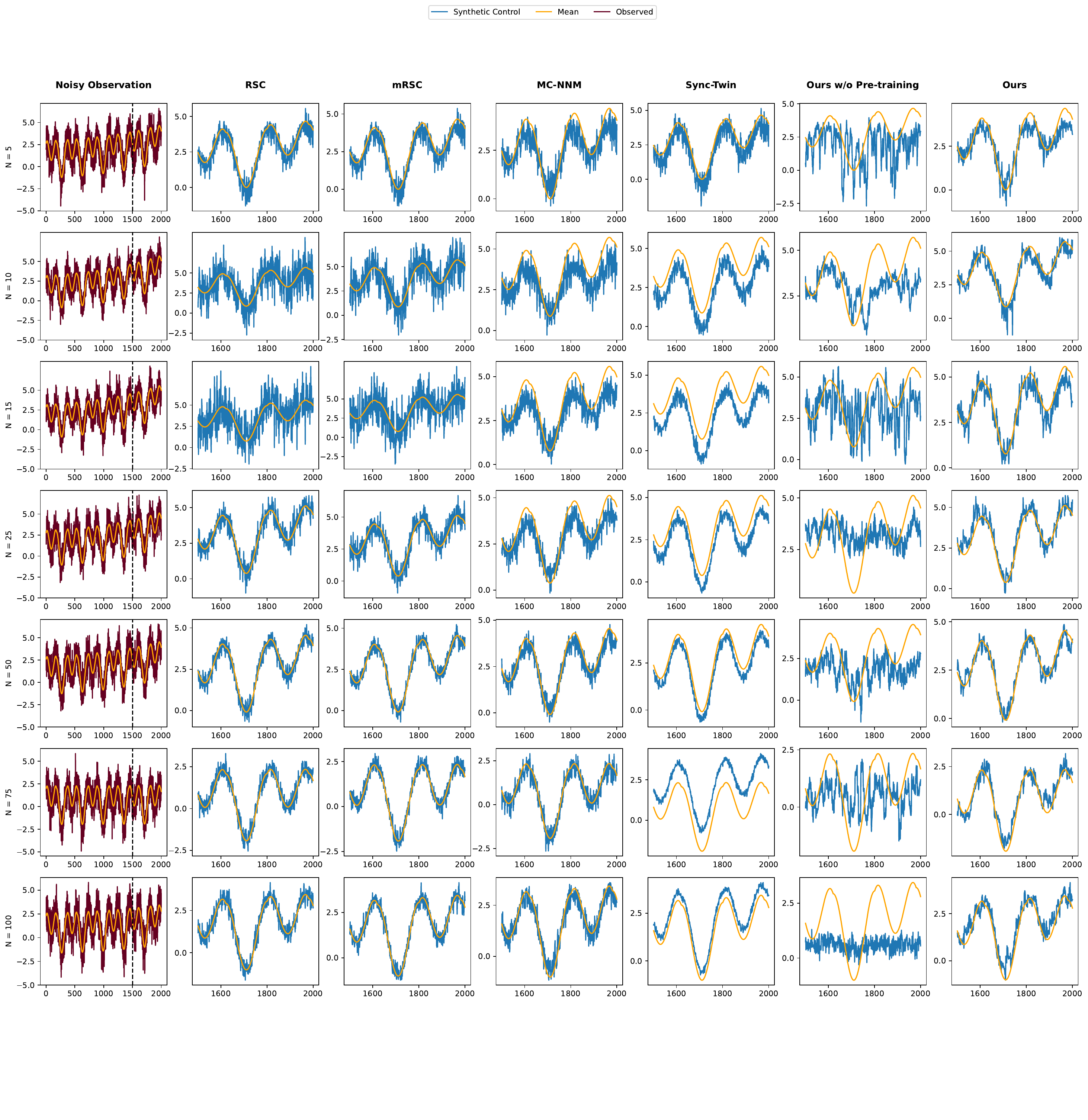}
    \caption{Effect of donor pool size on SC estimates. Our approach surpasses or performs competitively with prior estimators for all donor sizes in consideration}
    \label{fig:donors}
\end{figure*}

From Fig.~\ref{fig:donorpool}, we gain insight into the effect of the donor pool size on the accuracy of our prediction. Synthetic counterfactual problems often tend to have a small number of donors for eg. few states tend to experience similar policies, rare diseases inflict a small number of the population etc. Hence, we set $N \in [5, 10, 15, 25,50,75,100]$ with $\sigma^2=1$. Our estimator achieves a lower RMSE for most donor pool sizes with pre-training. We plot the SC estimates obtained by all the evaluated methods for different donor pool sizes in Fig.~\ref{fig:donors}.  Note that,
the performance without pre-training the model is worse and, therefore, this step is crucial for the model to be biased in the right way prior to finetuning. For $N=5$, we observe that our model underfits compared to other methods.  This is expected given the complexity of Transformer models and the data constraints for extremely few number of donors. However we outperform the other models as the number of donors increase. With the availability of more donors, the performance margin between linear 
estimators and our model narrows. We attribute this observation to two causes. First, we posit that the other models, which assume underlying linear factorized models with fixed temporal effects, need a higher number of donors to control for time-varying confounders and hence yield better predictions only when the donors increase. Second, the stronger choice of inductive bias by using all the donors simultaneously to predict the outcomes leads to slight overfitting for spatiotemporal transformers as the number of donors increase. This opens up interesting directions to prevent such overfitting via prior donor elimination, sparsity constraints on Transformer attention among other possible regularization methods and we leave this for future work.

Next, we evaluate our approach on real-life datasets at both population and individual levels spanning public
health, randomized trials, and rare diseases.

\subsection{Analysing Public Health Policy: California Proposition 99}
\begin{figure}[!htb]
    \centering
    \includegraphics[width=\linewidth]{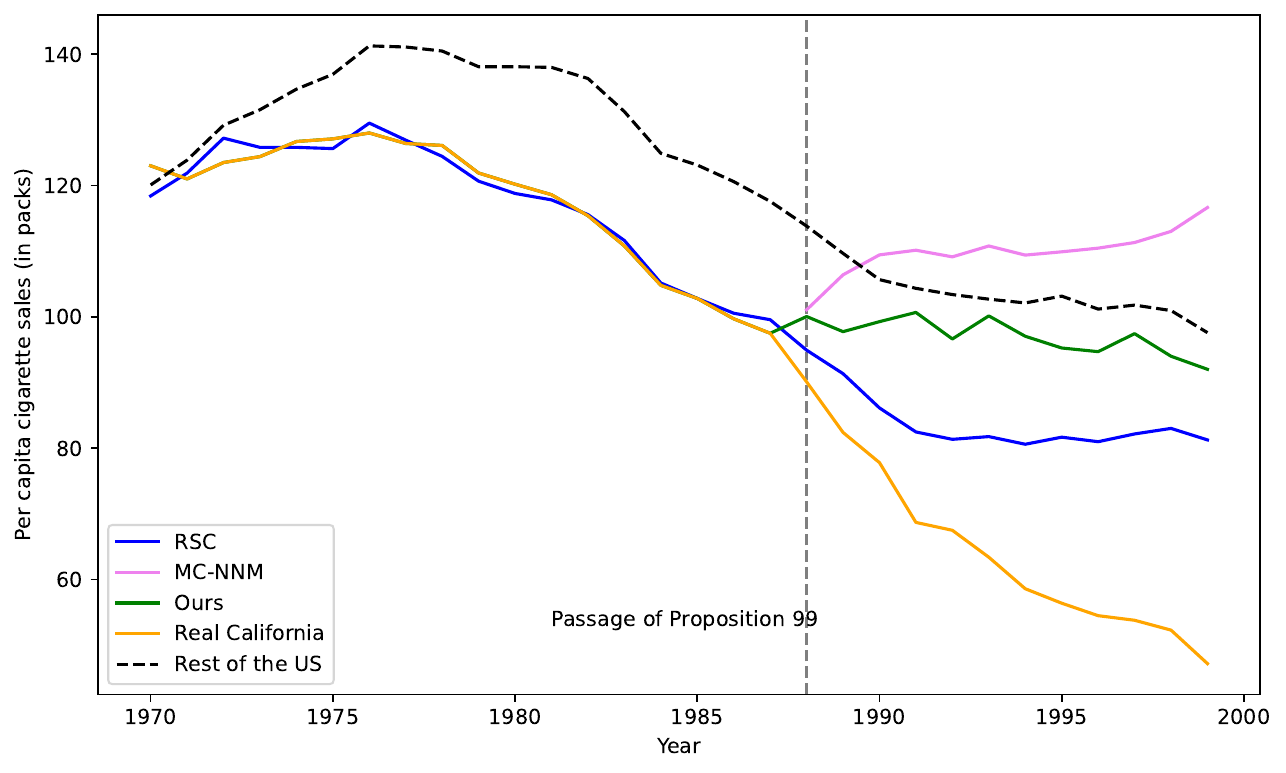}
    \caption{Comparison of methods on synthetic vs. true California. We infer that the passage of Proposition 99 
reduced per-capita cigarette sales from the gap between observed California (yellow) and the synthetic counterfactual of California (green). Our model also predicts that in the absence of Proposition 99, cigarette sales in California would slowly converge to the national average (dotted line). }
    \label{fig:prop99}
\end{figure}

\begin{figure}[!hbt]
\centering
\begin{subfigure}[t]{\textwidth}
\centering
    \includegraphics[width=0.8\textwidth]{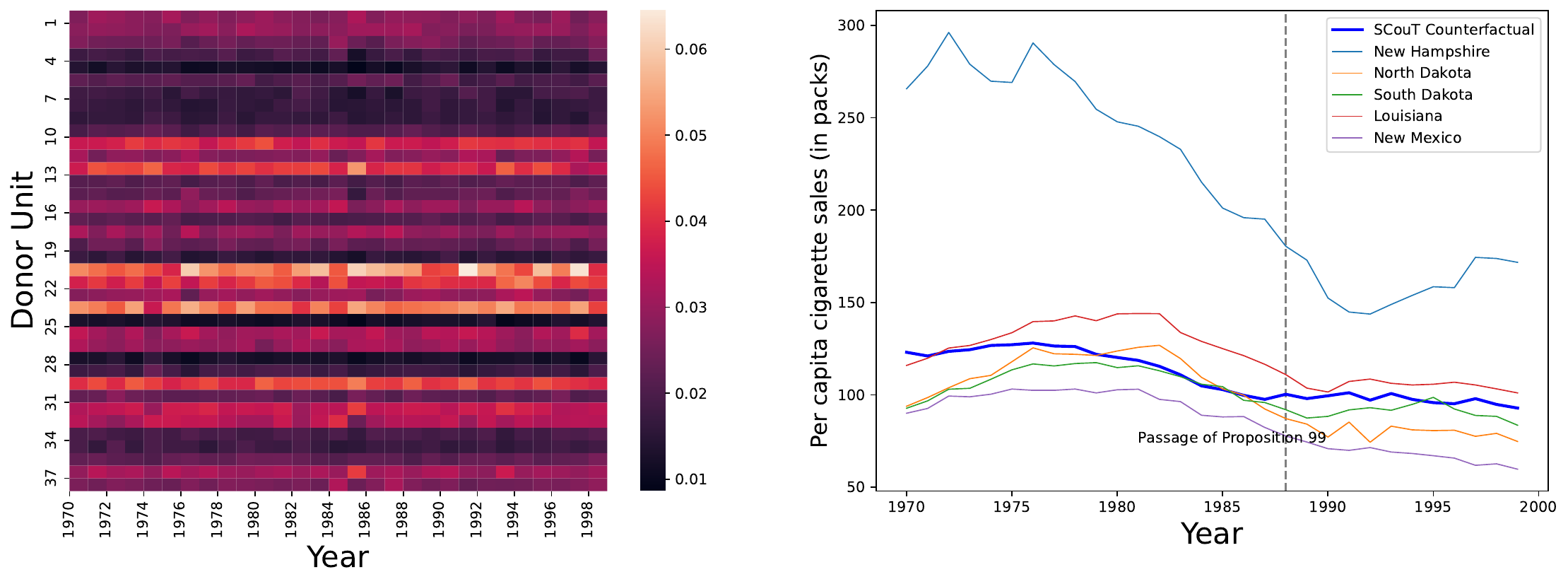}
    \caption{Layer 1 spatiotemporal donor attention}
    \label{fig:first_layer}
\end{subfigure}
\hfill
\begin{subfigure}[t]{\textwidth}
\centering
    \includegraphics[width=0.8\textwidth]{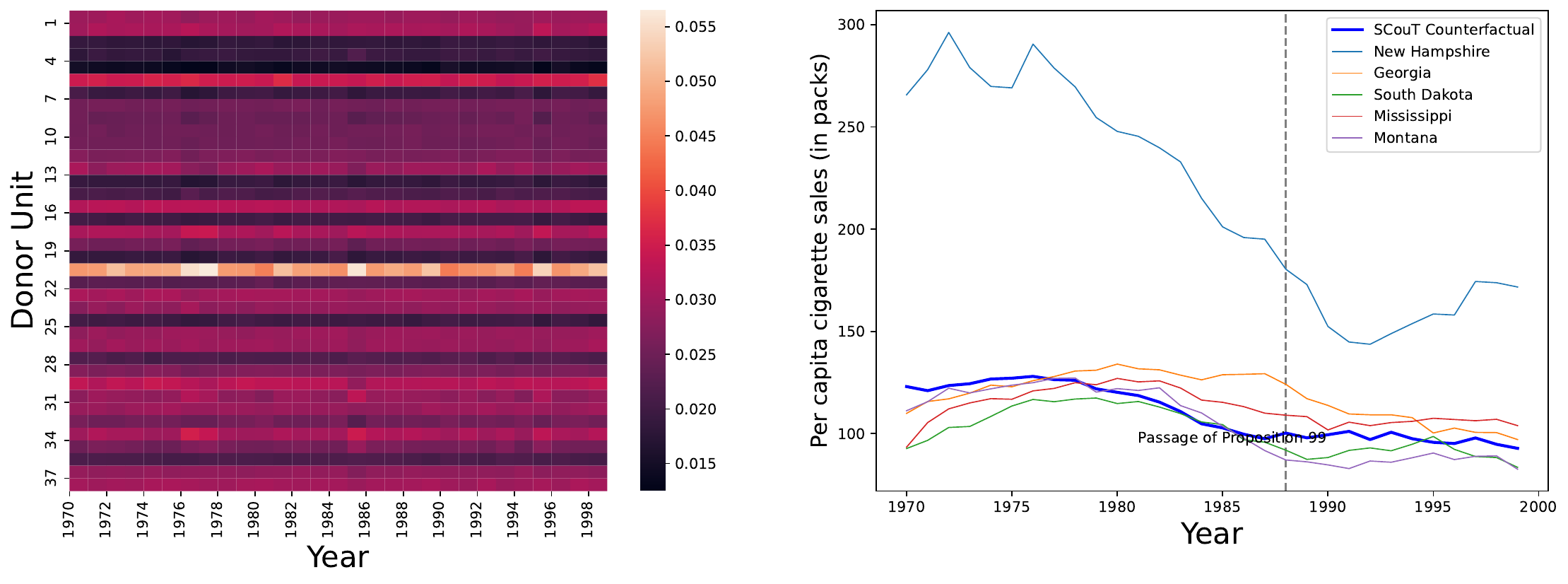}
    \caption{Layer 2 spatiotemporal donor attention}
    \label{fig:second}
\end{subfigure}
  
\begin{subfigure}[t]{\textwidth}
\centering
    \includegraphics[width=0.8\textwidth]{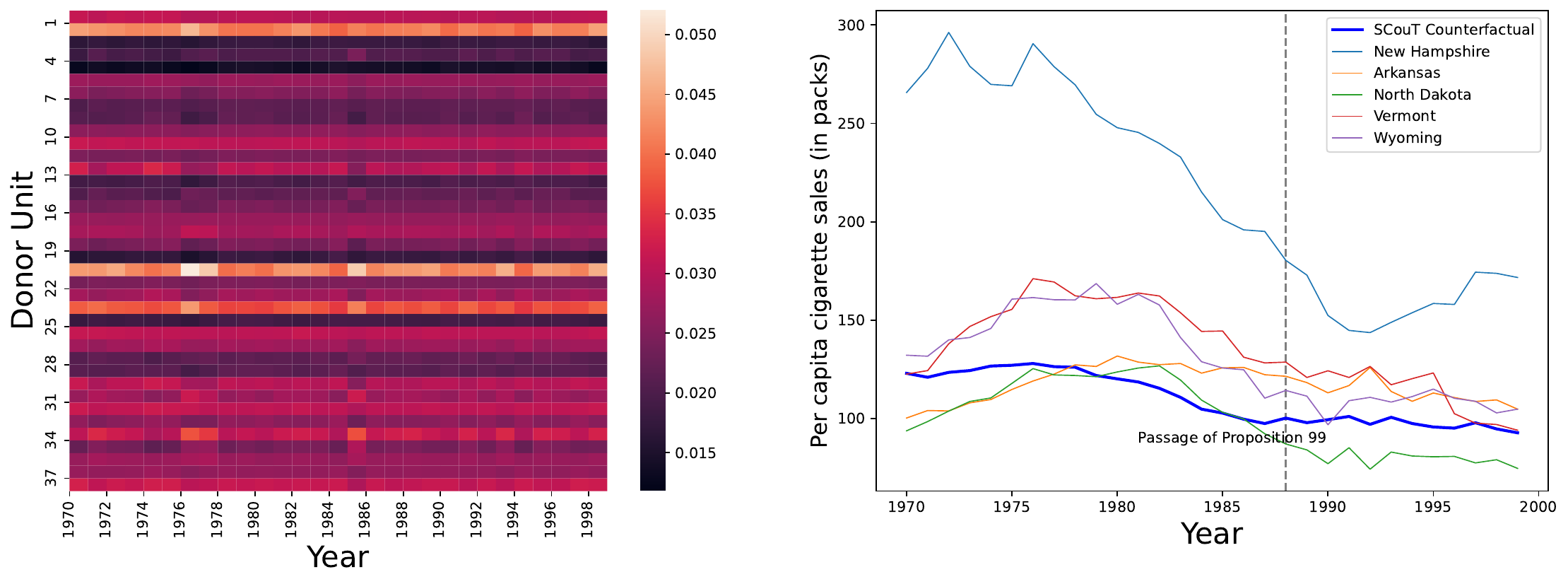}
    \caption{Layer 3 spatiotemporal donor attention}
    \label{fig:third}
\end{subfigure}
\hfill
\caption{Figures (a), (b), (c) on the left visualize donor attentions for the three layers of the Transformer. Attention is sparse and spread across specific donors, with the sparsity becoming more pronounced in the deeper layers. For each attention distribution map, we extract top 5 donors and plot them against our synthetic counterfactual on the right. We note that New Hampshire that has highest cigarette sale per capita among all the states is consistently used across all attention layers. On the other hand, the model selects specific donors at every layer to iteratively refine its counterfactual.}
\label{fig:prop99_weights}
\end{figure}

In 1988, California became the first in the United States to pass a large anti-tobacco law,
Proposition 99, that hiked the excise tax on cigarette sales by 25 cents. To evaluate the
effectiveness of this law, we construct a synthetic California without Proposition 99 and measure the
per capita cigarette sales in this synthetic unit. The donor pool consists of 38 control states where
no significant policies for tobacco control were introduced and additional covariates like beer
consumption, population, and income are included. Fig.~\ref{fig:prop99} compares the California
counterfactual prediction of our method and various baselines\footnote{mRSC yields a poor pre-intervention fit and has been omitted. Sync-Twin isn't amenable to missing outcome values and is omitted for real data evaluations.}, all showing that the cigarette sales would have been higher in the absence of the law. We infer that per-capita cigarette sales fell by 45 packets in real
California towards the year 2000 compared to synthetic California. Moreover, our model makes the intuitive prediction that in the absence of the law, cigarette sales in California would converge to the national average. As explained in Section \ref{sec:interpretability}, attention scores of the model can be used to extract the contribution of the donors in making the counterfactual prediction. These weights
across the donor space and time are illustrated in Fig.~\ref{fig:prop99_weights}, indicating the combination that best reproduces the outcome before the passage of Proposition 99.

\subsection{SCouT for Insilico Randomized Trails: Drug trials for Childhood Asthma}
\begin{figure}[!htb]
    \centering
    \includegraphics[width=\linewidth]{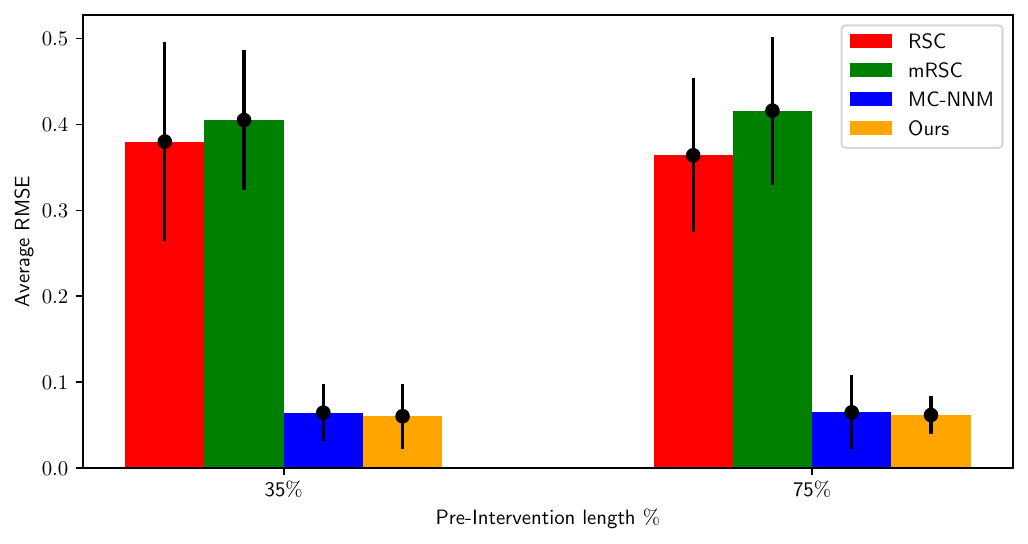}
    \caption{Average RMSE for post-intervention PreFF prediction across different pre-intervention lengths. Our method outperforms prior work on either pre-intervention lengths. Plotted at 90\% confidence intervals.}
    \label{fig:asthma_bar}
\end{figure}

\begin{figure}[!htb]
    \centering
    \includegraphics[width=0.7\linewidth]{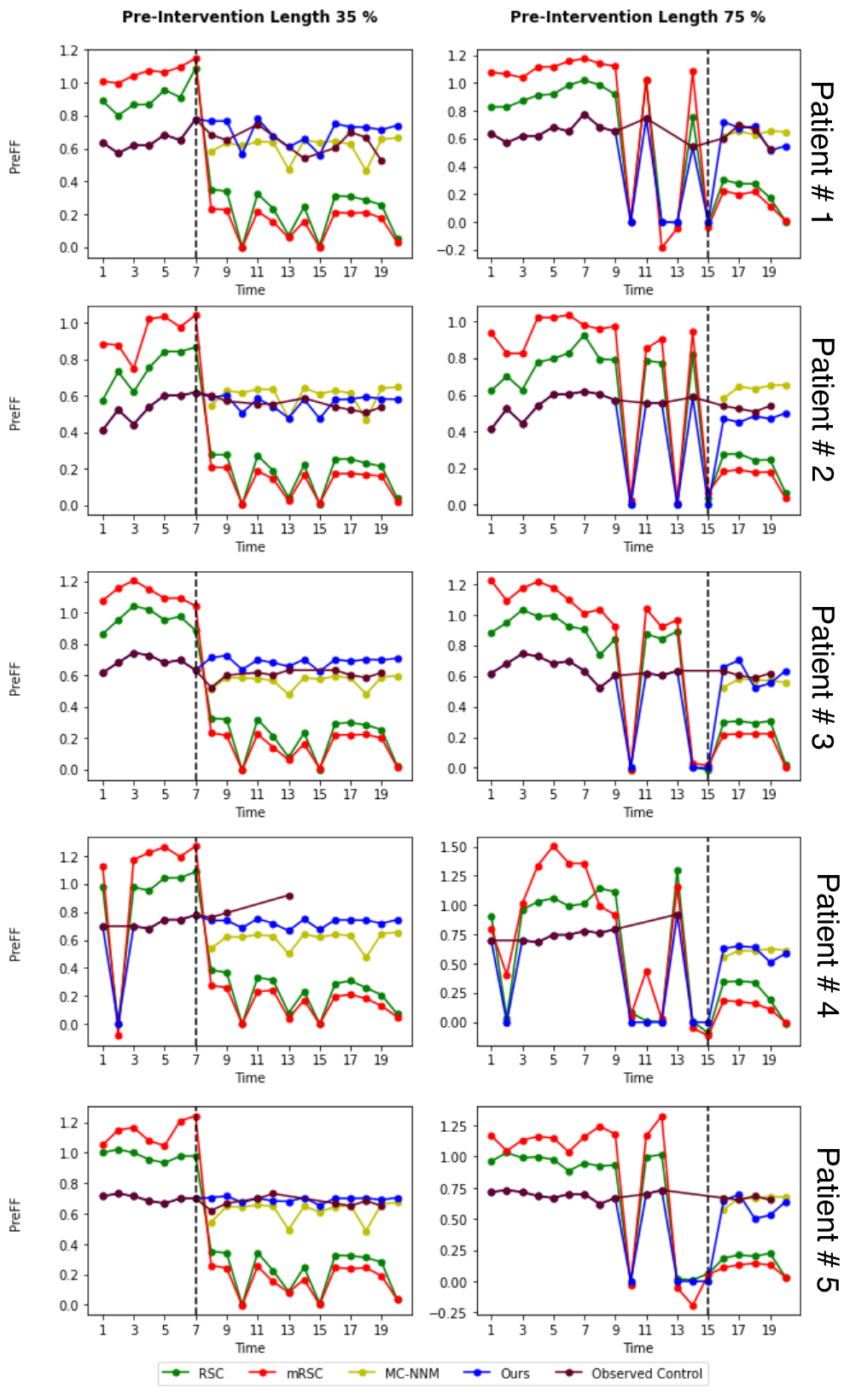}
    \caption{Estimates of PreFF for participants in the placebo group using other units in the placebo group as donors. The vertical black line indicates intervention instance. Our model reliably generates SC for both pre-intervention periods.}
    \label{fig:asthma}
\end{figure}

At the patient level, the synthetic counterfactual is a dynamic, virtual representation of the human being over
time and enables applications for in silico clinical trials. As an example, we look at The Childhood Asthma
Management Program (CAMP) \cite{camp}, an RCT designed to study the long-term pulmonary effects of three treatments
(budesonide, nedocromil, and placebo) on children with mild-to-moderate asthma. The trial's placebo arm contains anonymized longitudinal data of 275 patients with over 20 spirometry measurements per patient.  Pre-Bronchodilator Forced Expiratory Volume to Forced Vital Capacity ratio (PreFF) is a vital metric of lung capacity in Asthma patients that measures volume of air that an individual can exhale during a forced breath prior to the usage of a bronchodilator. Here, we  model the control arm of the RCT and predict the PreFF of a target patient using the other placebos as donors. We consider two settings, where pre-intervention lengths are 35\% and 75\% of the total trajectory. One at a time, we set one patient as the target unit and consider the others as donors. In this fashion, we model the control paths of the first five patients in the placebo arm, the ground truths available to us. The average RMSE across these patients 
is reported in Fig.~\ref{fig:asthma_bar}. We plot the counterfactual estimates for patients in Fig.~\ref{fig:asthma}. Our method generates a reliable control along with MC-NNM whereas estimates produced by RSC and mRSC
are highly biased. We posit that local spatiotemporal mapping is reasonably effective in controlling for time-dependent confounders, whereas linear estimators that assign time-agnostic weights suffer significantly.

\subsection{Me, My Doctor, and My Digital
Twin: A Case Study on Friedreich’s Ataxia}

\label{sec:fa}
FA is a fatal degenerative nervous system disorder with no cure. As a progressive disease with a wealth of available data and a
growing number of potential therapeutic interventions, FA is one of the many genetic diseases that can benefit from precision medicine. We use clinical data collected during the FA Clinical
Outcome Measure Study (FA-COMS) cohort \cite{Regner2012-kf}, a natural history study involving yearly
assessments of a core set of clinical measures and quality of life assessments, and maintained by the Critical Path Institute. 163 metrics were included in
our model based on clinical relevance and minimal missing data. Each metric can be classified as 1) either
longitudinal or static and 2) binary, ordinal, categorical, or continuous. Prior work only consider continuous data and unlike the generalized SCouT framework are therefore, not amenable to categorical data.  A recent study found that Calcitriol,
the active form of vitamin D, is able to increase Frataxin levels and restore mitochondrial function in cell
models of FA. FA is caused by a deficiency of Frataxin, Accordingly, Calcitriol supplements could potentially improve 
health outcomes for FA patients \cite{Britti2021-ux}. We explore the effect of Calcitriol as an example of a
synthetic medical intervention. Our resulting dataset tailored for the Calcitriol synthetic intervention
analysis included $N = 21$ donor units who indicated that they were taking a Calcitriol supplement at some
point during the study and had sufficient pre-intervention data for training (see Appendix \ref{sec:data_format} for data-formatting details). The dataset was aligned such that
every donor patient began taking the supplement at $T_0 = 8$. A few patients never indicated that they had taken
the supplement during the study period;
hence, we set these patients as the target unit and simulated the counterfactual situation under which the patients do receive treatment at $T_0$. It is important
to note that unlike randomized controlled trials, where we evaluate the effect of
a treatment on the average individual, the synthetic intervention presents the
predicted treatment effect on an individual patient, in line with the goals of precision
medicine. Imagine a FA patient walking into a clinic, where the physician has access to data from patient's previous health records and is deciding
whether or not to recommend a Calcitriol supplement. The physician can use the synthetic intervention technique to generate a counterfactual under this treatment for any desirable metric, as we demonstrate next. We evaluated the patients on several rating scales commonly used for FA patients \cite{Fahey2007iw}:
\begin{itemize}
    \item The FA Rating Scale neurologic examination (FARSn) \cite{Fahey2007iw}  is one of the most used tests to evaluate an FA patient’s disease progression. It involves scoring on several subscales broadly divided into bulbar, lower limb, upper limb, peripheral nervous system and upright stability functionalities.
    \item The Nine-Hole Peg Test (9-HPT) is a quantitative measure of finger dexterity and is conducted on both the dominant and non-dominant arm, in that order. The patient is told to pick up and place pegs into open holes on a board in front of them.
    \item The Activities of Daily Living (ADL) Scale is widely used to rank adequacy and independence in basic tasks that a person could expect to encounter every day, including grooming, dressing, walking, and drinking.
\end{itemize}

\begin{figure}[!htb]
    \centering
    \includegraphics[width=\linewidth]{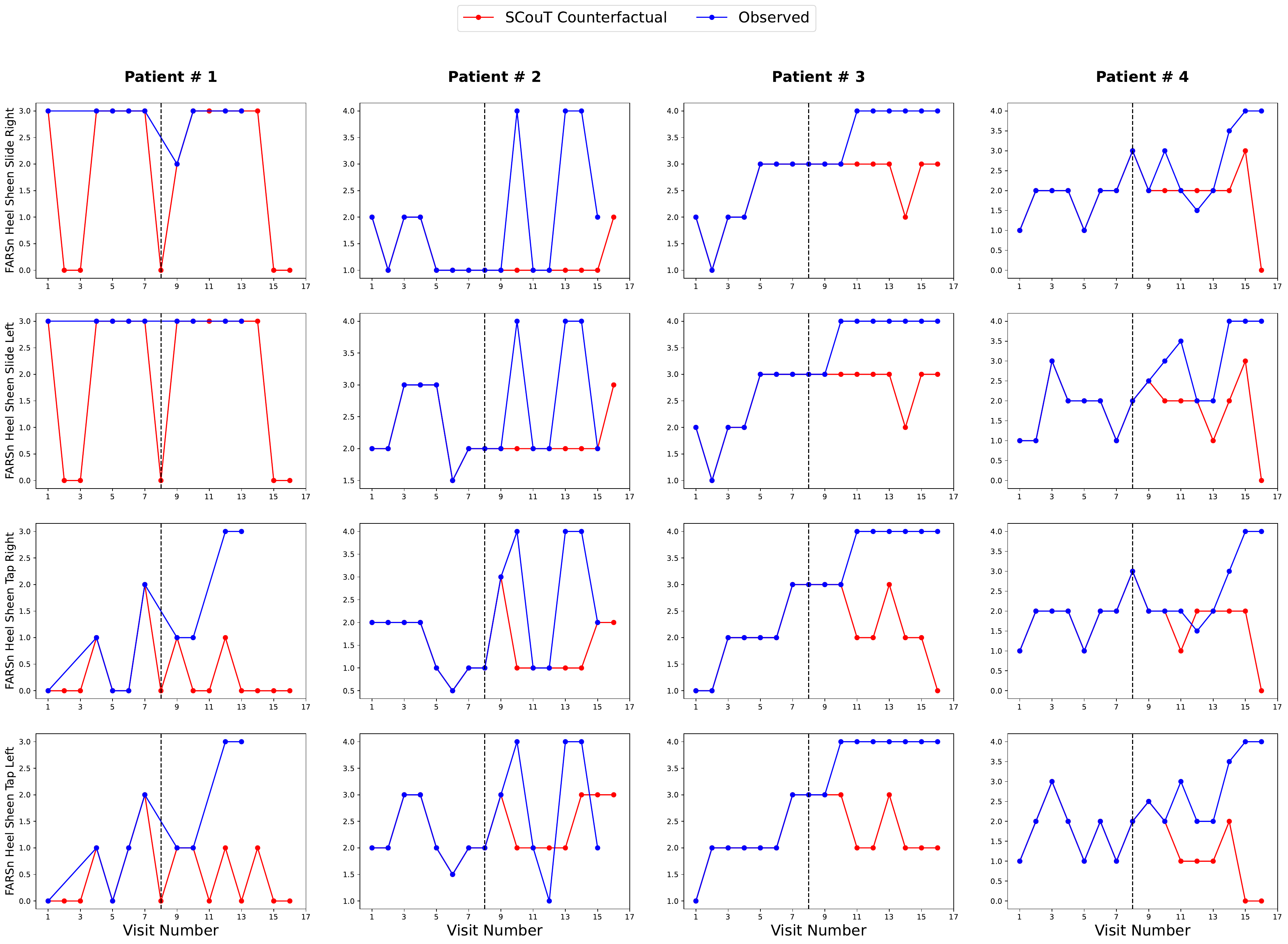}
    \caption{Tailored predictions for each patient under different FARSn scale metrics.
The vertical black line indicates the intervention instance. A higher score is associated with a greater degree of disability. Except for patient 2, SCouT predicts considerable improvements after using Calcitriol.}
    \label{fig:si_farsn}
\end{figure}

\begin{figure}[!htb]
    \centering
    \includegraphics[width=0.8\linewidth]{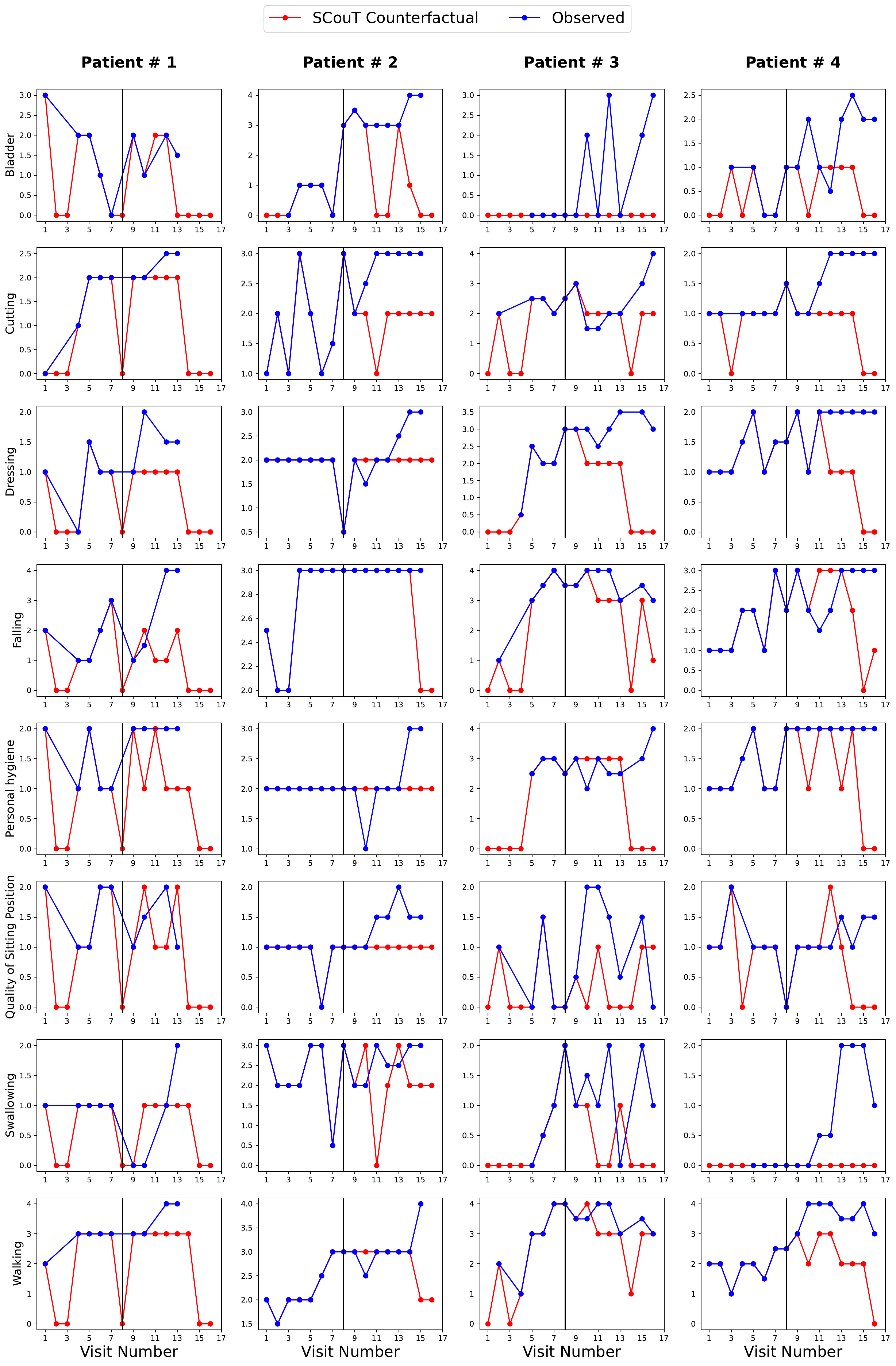}
    \caption{Tailored predicted scores for each patient under different ADL scale metrics. The vertical black line indicates intervention instance. A higher score is associated with a greater degree of disability. SCouT predicts considerable improvements after using Calcitriol across all metrics and all patients.}
    \label{fig:si_adl}
\end{figure}

\begin{figure}[!htb]
    \centering
    \includegraphics[width=\linewidth]{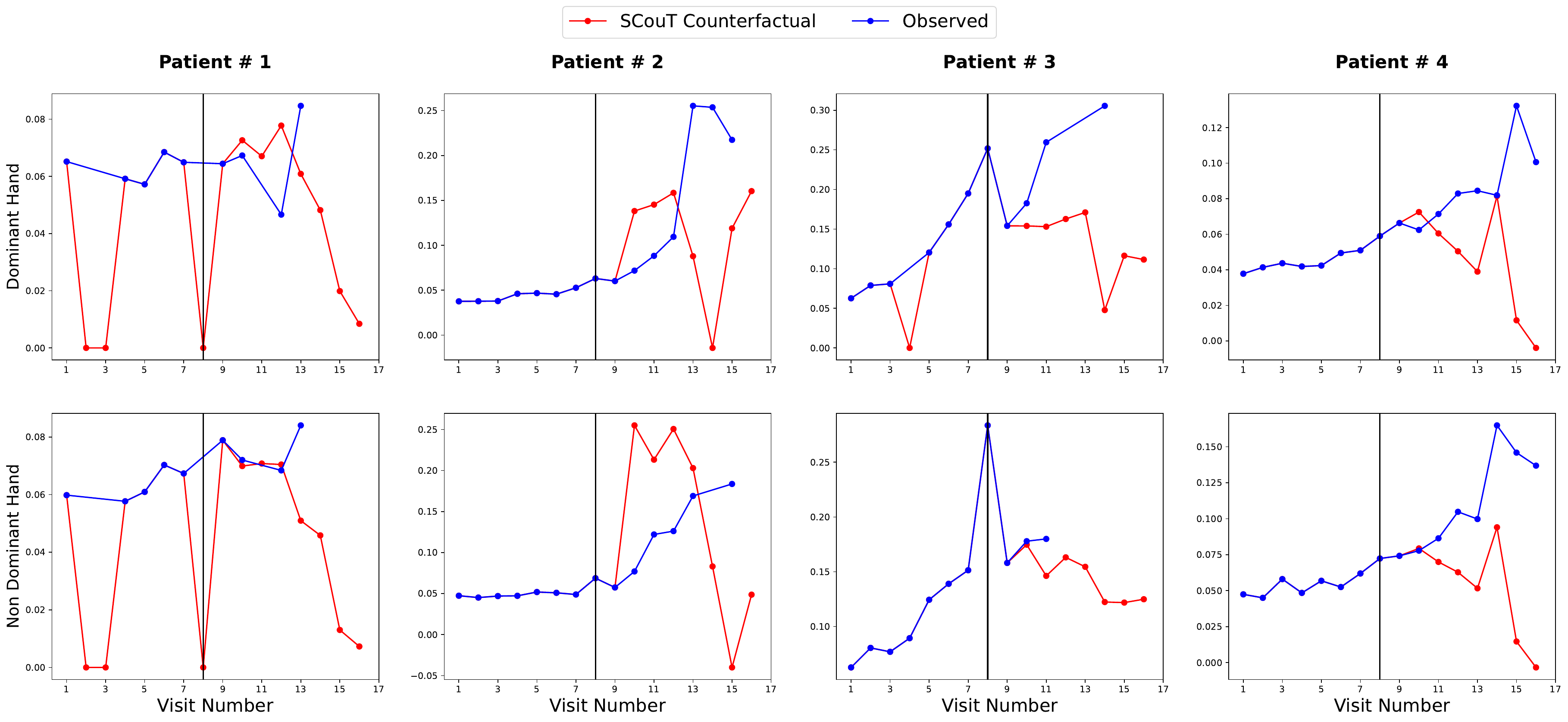}
    \caption{Tailored counterfactuals for each patient under 9-HPT metrics. The vertical black line indicates the intervention instance. The scores measure the time it takes for the patient to place the pegs into the hole; a higher score indicates a greater level of disability. SCouT predicts improvement in test times for all patients across both the dominant and non-dominant hands.}
    \label{fig:si_hpt}
\end{figure}

We compare the predictions under the counterfactual scenario where the patients begin to take a
Calcitriol supplement at time $T_0$ against the true observed values for the patients with no intervention. We
included metrics from the FARSn, 9-HPT, and ADL exams that demonstrated the counterfactual under Calcitriol use for the patients. Fig. \ref{fig:si_farsn}, Fig. \ref{fig:si_adl}, Fig.\ref{fig:si_hpt} show the predictions for the FARSn, ADL, and 9-HPT scales, respectively. Most counterfactuals seem to suggest that Calcitriol use would have been beneficial with respect to all metrics highlighted below.

Across specific metrics and select patients, comparing the Calcitriol counterfactual and observed patient values also predicted
little benefit from the supplement or were otherwise difficult to extract qualitative meaning from. Finally,
none of the comparisons indicated that Calcitriol would allow further disease progression in the patients, 
which is comforting evidence that, at worst, Calcitriol has little effect. At best, the supplement could greatly improve the quality of life for a patient with FA.


\section{Conclusion}
SC-based methods paved the way for more effective techniques to estimate the
counterfactual of a unit from several donor units. However, these methods are still
restricted because they cannot capture inter-unit and intra-unit temporal contexts and are parametrically constrained to model linear dynamics. We introduced a 
Transformer-based Seq2Seq model that leverages local spatiotemporal relations to estimate the
counterfactual sequence through multimodal inputs better and overcome these limitations. Experiments on synthetic and real-world data
demonstrate that our model has high predictive power even in noisy data and donor pools of varying sizes. Explicitly modeling donor units is a strong inductive bias and significantly enhances predictive power compared to time series modeling. Our method leverages the Transformer model's credit assignment ability to analyse the
first-of-its-kind temporal evolution of donor contribution. We also demonstrated the applicability of our
Transformer-based model in simulating counterfactual scenarios in healthcare, e.g., to
accelerate in-silico clinical trials by simulating the control counterfactual for individuals in an Asthma
randomized trial and generating the Calcitriol intervention counterfactual for FA patients. As big data and artificial intelligence continue to revolutionize healthcare, our work can be adapted for use beyond clinical trials and clinical decision-making and create measurable improvements in public health.

\section{Acknowledgement}

This work was supported by NSF Grant No. CCF-2203399. The experiments reported in this paper were performed on the computational resources
managed and supported by Princeton Research Computing at Princeton University.

\bibliography{references}
\bibliographystyle{IEEEtran.bst}
\appendix

\appendix

\section{Model Details}

Our model is implemented using the Hugging Face library \cite{wolf2020huggingfaces}. Our choice of the encoder-decoder model is a Bert2Bert model. The hyperparameters we use for various experiments are outlined in Table \ref{tab:hyperparams}.

\begin{table}
\caption{Model and training hyperparameters}
\label{tab:hyperparams}
\vskip 0.15in
\begin{center}
\begin{small}
\begin{sc}
\begin{tabular}{cc}
\toprule
Hyperparameter  & Value\\
\midrule
Number of Layers &  \thead{Synthetic: 3\\Prop. 99: 3\\Asthma: 4\\FA: 4}\\
Number of Attention Heads & 1 \\ 
Hidden Dimension & \thead{Synthetic: 32\\Prop. 99: 128\\Asthma: 128\\FA: 128} \\
Non linearity & GELU \\
Dropout & 0.1\\
Pre-Intervention Length & \thead{Synthetic:\\ 20 (N=5,10,15)\\10 (N=25,50,75,100)\\ Prop. 99: 5\\Asthma: 2\\FA: 5}\\
Post-Intervention Length & \thead{Synthetic:\\ 10 (N=5,10,15)\\5 (N=25,50,75,100)\\ Prop. 99: 2\\Asthma: 2\\FA: 2}\\

\#Singular Values Retained & \thead{Synthetic: 5 \\ Prop. 99: 3 \\Asthma: 55\\FA: 5}\\
Learning Rate & $1 \times 10^{-4}$  \\
Weight Decay & $1 \times 10^{-4}$ \\

Warmup Steps (Pre-training) & \thead{Synthetic: $1 \times 10^{4}$ \\Prop. 99: $1 \times 10^{3}$ \\Asthma: $1 \times 10^{4}$ \\ FA: $1 \times 10^{4}$ }\\
Pre-training Iterations & \thead{Synthetic: $5 \times 10^{4}$\\Prop. 99: $5 \times 10^{3}$\\Asthma: $5 \times 10^{4}$\\FA: $5 \times 10^{4}$}\\
Fine-tuning Iterations & \thead{Synthetic: $1 \times 10^{4}$\\Prop. 99: $1 \times 10^{3}$ \\Asthma: $5 \times 10^{3}$\\FA: $5 \times 10^{4}$}\\
Batch Size  & \thead{Synthetic: 128\\Prop. 99: 64\\Asthma: 16 \\ FA: 16} \\
\bottomrule
\end{tabular}
\end{sc}
\end{small}
\end{center}
\vskip -0.1in
\end{table}

\section{Proposition 99}
Five covariates are used for generating the control:
\begin{itemize}
    \item Per-capita cigarette consumption (in packs)
    \item Per-capita state personal income (logged)
    \item Per-capita beer consumption
    \item Percent of state population aged 15–24
    \item Average retail price per pack of cigarettes (in cents)
\end{itemize}

\section{Asthma}
The CAMP dataset uses the Forced Expiratory Volume (FEV) and Forced Vital Capacity (FVC) as the lung functions.  16 continuous covariates from the dataset are used for generating the control:
\begin{itemize}
    \item Pre-Bronchodilator FEV/FVC ratio \% (PREFF) 
    \item Age in years at randomization (age rz)
    \item Hemoglobin (hemog)
    \item Pre-Bronchodilator FEV (PREFEV) 
    \item Pre-Bronchodilator FVC (PREFVC)  
    \item Pre-Bronchodilator peak flow (PREPF)
    \item Post-Bronchodilator FEV (POSFEV)
    \item Post-Bronchodilator FVC (POSFVC)
    \item Post-Bronchodilator FEV/FVC ratio \% (POSFF)
    \item Post-Bronchodilator peak flow (POSPF)
    \item Pre-Bronchodilator FEV \% prediction (PREFEVPP)
    \item Pre-Bronchodilator FVC \% prediction (PREFVCPP)
    \item Post-Bronchodilator FEV \% prediction (POSFEVPP)
    \item Post-Bronchodilator FVC \% prediction (POSFVCPP)
    \item White blood cell count (wbc) 
    \item Age of current home (agehome)
\end{itemize}

\section{Friedreich’s Ataxia}

\subsection{FA Metrics}
The following metrics were included in the clinical dataset for FA:

\begin{itemize}
    \item Age of onset
    \item Length of shorter GAA Repeat
    \item Presence of point mutation
    \item Death Flag
    \item Ethnicity
    \item Sex
    \item Race
    \item Assistive Device Type 
    \item Level of education
    \item Living circumstances
    \item The Friedreich ataxia Rating Scale neurologic (FARSn) examination subscores 
    \item The Timed 25-Foot Walk 
    \item The Nine-Hole Peg Test: Dominant Hand and Non-Dominant Hand
    \item Activities of Daily Living (ADL) Scale
    \item Bladder Control Scale
    \item Bowel Control Scale
    \item Impact of Visual Impairment Scale
    \item Modified Fatigue Impact Scale
    \item MOS Pain Effects Scale
    \item  SF-10 and SF-36
\end{itemize}

\subsection{Data Formatting for FA}
\label{sec:data_format}

\begin{figure}[!hbt]
    \centering
    \includegraphics[width=0.8\textwidth]{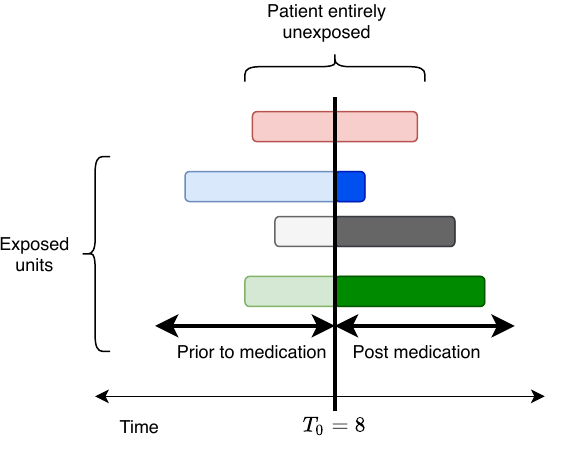}
    \caption{Dataset formatting to align all exposed units at a common intervention instance.}
    \label{fig:fa_dataset}
\end{figure}

Since the FA dataset we use comes from a natural history study that does not introduce any interventions itself, we have to reformat the dataset so it can be used for a synthetic intervention. The FA data includes longitudinal information regarding the medications a patient is taking. Out of all the patients, 72 indicated that they were
taking a Calcitriol supplement at some point during the study. We manually searched for patients that reported taking Calcitriol for an extended period of time (typically 3 or more years) and had sufficient pre-intervention data for training. The resulting dataset tailored for the Calcitriol synthetic intervention analysis included $N = 21$ and was aligned using the methodology from Figure \ref{fig:fa_dataset}. Thus every donor patient began taking the supplement at $T_0 = 8$. Four patients who had not taken the supplement during the study period, were then set as the target unit. Furthermore, the narrowed scope of the question we investigate in this section also gives us a use case to test our model on real world FA data with a much smaller number of donor units.

\end{document}